\pdfoutput=1
\documentclass[11pt]{article}

\usepackage[preprint]{acl}

\usepackage{times}
\usepackage{latexsym}

\usepackage[T1]{fontenc}
\usepackage[utf8]{inputenc}

\usepackage{microtype}

\usepackage{inconsolata}

\usepackage{graphicx}
\usepackage{booktabs}
\usepackage{array}
\usepackage{makecell}
\usepackage{colortbl}
\usepackage{xcolor}
\usepackage{todonotes}
\usepackage{listings}
\usepackage{tabularx}
\usepackage{multicol}
\usepackage{amsmath}
\usepackage{multirow}
\usepackage{comment}
\usepackage{svg}
\usepackage{float}

\definecolor{codered}{rgb}{0.9,0,0}

\title{Sketch2Code: Evaluating Vision-Language Models for Interactive Web Design Prototyping}

\author{Ryan Li \\
  Stanford University \\
  \texttt{lansong@stanford.edu} \\\And
  Yanzhe Zhang \\
  Georgia Tech \\
  \texttt{z\_yanzhe@gatech.edu} \\\And
  Diyi Yang\\
  Stanford University \\
  \texttt{diyiy@stanford.edu} \\
  }

\begin{document}
\maketitle
\begin{abstract}
% Foundation models have demonstrated significant potential to automate development workflows, as visual understanding and code generation capabilities enable them to turn visual designs into implementation end-to-end.
%Sketch is a widely used, intuitive tool for early-stage web design, allowing designers to quickly communicate and visualize ideas. However, current design automating benchmarks focus solely on transforming high-fidelity inputs instead of sketches into code implementations. 
Sketches are a natural and accessible medium for UI designers to conceptualize early-stage ideas. However, existing research on UI/UX automation often requires high-fidelity inputs like Figma designs or detailed screenshots, limiting accessibility and impeding efficient design iteration.
To bridge this gap, we introduce Sketch2Code, a benchmark that evaluates state-of-the-art Vision Language Models (VLMs) on automating the conversion of rudimentary sketches into webpage prototypes. Beyond end-to-end benchmarking, Sketch2Code supports interactive agent evaluation that  %  Unlike previous efforts, Sketch2Code is a multi-turn interactive framework that 
mimics real-world design workflows, where a VLM-based agent iteratively refines its generations by communicating with a simulated user, either passively receiving feedback instructions or proactively asking clarification questions. We comprehensively analyze ten commercial and open-source models, showing that Sketch2Code is challenging for existing VLMs; even the most capable models struggle to accurately interpret sketches and formulate effective questions that lead to steady improvement.
% An average commercial model improves only 1.6\% visual similarity and 0.8\% layout similarity over five rounds of interactions. 
Nevertheless, a user study with UI/UX experts reveals a significant preference for proactive question-asking over passive feedback reception, highlighting the need to develop more effective paradigms for multi-turn conversational agents \footnote{Code/Data available on project Page: \url{https://salt-nlp.github.io/Sketch2Code-Project-Page/}}.
\end{abstract}

\section{Introduction}

% \ryan{perhaps add more context about frontend ui designs}
Large Language Models (LLMs) have spurred a variety of applications on automating functional code implementations from natural language instructions \citep{Le_2020, chen2021evaluatinglargelanguagemodels, li2023starcodersourceyou, jimenez2024swebenchlanguagemodelsresolve}. Recent works
% few studies have focused on frontend User Interface (UI) applications due to challenges in processing multimodal inputs and vast search spaces.
such as \citet{si2024design2code} and \citet{laurençon2024unlockingconversionwebscreenshots} have started to explore possibilities of generating HTML code directly from full-fidelity web designs (e.g, mock-up screenshots) using Vision Language Models (VLMs), aiming to democratize frontend design for researchers, practitioners, and general users.
However, the screenshot-to-code setting is inconvenient as providing detailed graphical designs for the desired User Interface (UI) is time-consuming and sometimes requires professional tools with a steep learning curve. 
% prohibiting regular users from implementing their custom web UIs and making it inconvenient for UI designers to flesh out, communicate, and evaluate early-stage ideas. 
On the other hand, sketching is a low-fidelity, accessible, and plentiful tool that is much easier to learn and implement \citep{10.1145/3613905.3636266}. Despite their low fidelity, sketches are commonly used to ideate, communicate, and visualize design concepts, often serving as the earliest yet vitally important step of UI designs \citep{buxton2010sketching, bao2018interplay, 10.1145/3544549.3574187, 10.1145/3613905.3636266}. 
% \ryan{I'm not sure how to properly cite this in the appendix.}

Transforming sketches to code used to be implemented in a pipeline fashion that involves pattern and object recognition \citep{azure_2018, robinson2019sketch2codegeneratingwebsitepaper, jain2019sketch2codetransformationsketchesui, baulé2021automaticcodegenerationsketches}.
% In 2018, Microsoft launched an early attempt at converting whiteboard sketches to code implementations\footnote{https://azure.microsoft.com/en-us/blog/turn-your-whiteboard-sketches-to-working-code-in-seconds-with-sketch2code/}. \ryan{there doesn't seem to be a paper for this to cite.} While several works came out along the same direction in the following years \citep{robinson2019sketch2codegeneratingwebsitepaper, jain2019sketch2codetransformationsketchesui, baulé2021automaticcodegenerationsketches}, all of them are constrained to simple pattern and object recognition with hard-coded (Limited?) UI components.
However, recent development of general-purpose VLMs \citep{alayrac2022flamingovisuallanguagemodel, liu2023visualinstructiontuning, 2023GPT4VisionSC, reid2024gemini} began to shift this paradigm by enabling such transformation end-to-end. In this paper, we present \textbf{Sketch2Code}, a first-of-its-kind framework to access VLMs' capability of implementing web UI from user sketches, where we (1) collected 731 high-quality sketches from 484 real-world webpages through crowd workers based on \citet{si2024design2code}, (2) assessed VLMs' performance on directly transforming sketches to code, and (3) designed a multi-turn, interactive framework to benchmark VLMs on Sketch2Code using LLM simulated users, unlike prior works \citep{si2024design2code, laurençon2024unlockingconversionwebscreenshots} that focused solely on single-turn generations.

\begin{figure*}[t]
\centering
\includegraphics[width=\textwidth]{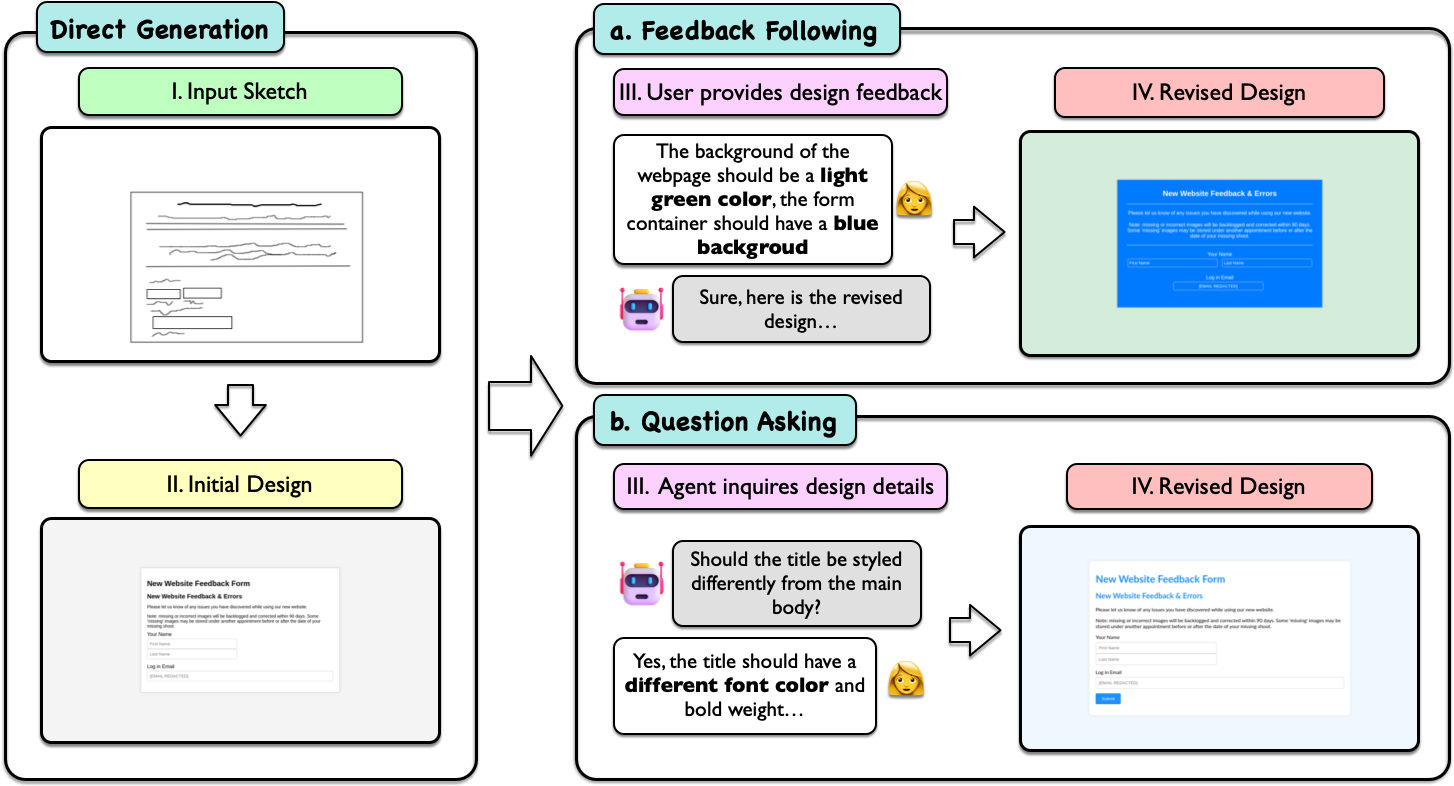}
\caption{Benchmark Overview. We provide an example of direct generation on the left. On the right, we show two examples of user-agent interactions in multi-turn scenarios: feedback following and question answering.}
\vspace{-1em}
\label{fig:benchmark_examples}
\end{figure*}

Real-world web design is an iterative process where initial concepts undergo multiple revisions based on continuous feedback and clarifications \citep{wynn2017perspectives}. Especially while using sketches, which are low-fidelity representations, it is impossible to figure out specific details, such as stylistic information, without additional input from the user.
%, hence making the Sketch2Code task unsuitable for single-turn setups. \ryan{Can someone double check if the writing here makes sense?}
To mirror realistic design workflows and study how well VLMs can interact with humans, our framework further introduces two multi-turn evaluation scenarios between a sketch2code agent and a simulated user: (1) the sketch2code agent follows feedback from the simulated user (\textbf{feedback following}) and (2) the sketch2code agent proactively asks the simulated user questions for design details and clarification (\textbf{question asking}). To this end, our framework assesses not only the ability of models to generate initial implementations based on abstract inputs but also their capacity to adapt and evolve these implementations in response to user feedback.
Since these are two of the most common communication patterns between human collaborators, our framework allows the simulated user to be seamlessly replaced by real users in a real-world deployment. Human annotations reveal that the simulated user provides faithful and meaningful feedback 83.3\% of the time and answers 86.7\% of the questions accurately.
% \yanzhe{probably add more clarification here.}

% Sketch2code not only extends the boundaries of current AI capabilities but also serves as a valuable testbed for assessing VLMs' understanding of high-level visual structures and their ability to interact iteratively with users to refine ambiguous inputs into precise implementations.

\noindent
We conducted a comprehensive analysis of ten models: GPT-4o, GPT-4o mini, Gemini 1.5 Pro, Gemini 1.5 Flash, Claude 3.5 Sonnet, Claude 3 Opus/Sonnet/Haiku, Llava-1.6-8b, and InternVL2-8b. Results indicate that inferring the correct layout structures from rudimentary sketch designs is challenging for VLMs in a single turn (\S~\ref{sec:results_direct}). While commercial models perform reasonably well in following user feedback—achieving performance improvements of up to 7.1\% in visual similarity \citep{si2024design2code} and 2.7\% in IoU-based \citep{rezatofighi2019generalized} layout similarity within five rounds of interaction—all VLMs struggle to formulate meaningful questions about the sketches and fail to reliably enhance their performance across multiple rounds in the question-asking scenario (\S~\ref{sec:results_multiturn}).

\noindent
To understand the utility of our sketch-based, multi-turn framework, we conducted a user study involving eight UI/UX practitioners while all participants recognized the usefulness of our framework. Furthermore, it reveals that the question-asking mode is \textbf{significantly preferred} over the more traditional feedback-following mode despite current suboptimal performances. Specifically, users prefer the agent to proactively undertake more of the cognitive workload and guide the design choices via a series of targeted questions.
% Specifically, the agent is expected to proactively undertake more of the cognitive workload, guiding users through design choices via a series of targeted questions.
Although questions asked by agents are often found ineffective, replacing model-generated questions with questions asked by human experts leads to a significant improvement with each question (\S~\ref{sec:user_study}).
Our findings highlight a critical gap between user expectations and current model capabilities, necessitating further research into human-AI collaboration and the capability of more proactive interactions.

\section{Related Work}
% \paragraph{Vision Language Models.} VLMs extend the success of LLMs to the multimodal domain. Blip 2 \citep{li2023blip2} connects Vision Transformers \citep{dosovitskiy2021imageworth16x16words} with large language models using Q-Former, while models like LLaVA \citep{liu2023visualinstructiontuning} and InstructBLIP \citep{dai2023instructblipgeneralpurposevisionlanguagemodels} use instruction tuning to improve generalization. While commercial Vision Language Models, such as GPT-4V \citep{2023GPT4VisionSC}, Claude3 \citep{anthropic2024claude}, and Google Gemini \citep{reid2024gemini}, have lately shown great success and achieve SoTA results in a wide range of tasks, innovations have also been made in the open-source community. OpenVLA \citep{kim2024openvlaopensourcevisionlanguageactionmodel} focuses on robotic manipulation, integrating DINOv2 \citep{oquab2024dinov2learningrobustvisual} and SigLIP \citep{zhai2023sigmoidlosslanguageimage} vision backbones with the Llama 2 \citep{touvron2023llama2openfoundation} language model. Yi \citep{ai2024yiopenfoundationmodels} improves the model scalability and generalizability using a cascaded data deduplication and quality filtering pipeline. InternVL2 \citep{chen2024far} achieves results comparable to SoTA commercial models across multiple tasks by using a continue learning strategy on vision encoder, dynamic high-resolution, and high-quality bilingual dataset.

\paragraph{LLM-Based Code Generation.} LLMs designed explicitly for coding, such as Codex \citep{chen2021evaluatinglargelanguagemodels}, StarCoder \citep{li2023starcodersourceyou}, InCoder \citep{fried2023incodergenerativemodelcode}, CodeLlama \citep{rozière2024codellamaopenfoundation}, and DeepSeek-Coder \citep{guo2024deepseekcoderlargelanguagemodel}, facilitate programming support applications like automatic code completion and infilling, as well as enabling users to interact with codebases. For general-purpose LLMs, adding code into the pretraining data also improves reasoning \citep{ma2023trainingcode, zhang2024unveilingcode}. However, the trend of evaluating coding ability on problem-solving benchmarks like HumanEval \citep{chen2021evaluatinglargelanguagemodels} and MBPP \citep{austin2021mbpp} leads to overlooking other realistic coding tasks, such as writing code for solving GitHub issues \citep{jimenez2023swebench} and, in our case, implementing an HTML website.

The introduction of vision modality further poses challenges to the task of code generation. Most of the open vision-language data focus on open-ended visual questions answering, thus essentially limiting the coding capability of open VLMs \citep{li2023blip2, liu2023visualinstructiontuning, dai2023instructblipgeneralpurposevisionlanguagemodels}, while commercial VLMs, such as GPT-4V \citep{2023GPT4VisionSC}, Claude3 \citep{anthropic2024claude}, and Google Gemini \citep{reid2024gemini} achieves remarkable progress probably due to more diverse and larger scale data collection. In this work, we find that open VLMs like InternVL2 \citep{chen2024far}, which achieves results comparable to commercial models across popular benchmarks, still lags far behind in terms of code generation and multi-turn interaction.
% This advancement prompts new HCI studies focused on designing enhanced programming tools for improved human-AI collaboration \citep{vasconcelos2023generationprobabilitiesenoughexploring, liang2024largescale}.

\paragraph{Frontend UI Code Generation.} \citet{7372013} pioneered reverse engineering mobile UIs using OCR and computer vision techniques to generate code. Pix2Code \citep{beltramelli2017pix2codegeneratingcodegraphical} introduced an end-to-end UI-to-code system leveraging CNNs and RNNs, but faced challenges with complex visual encoding and text decoding. \citet{8741736} incorporated neural network-based object detection and semantic segmentation into this process. Prior studies have also attempted automatic UI generation from sketches, such as \citet{azure_2018, robinson2019sketch2codegeneratingwebsitepaper, jain2019sketch2codetransformationsketchesui, baulé2021automaticcodegenerationsketches}, but are limited to simple pattern matching and object detection, with limited support in HTML syntax. Recently, \citet{soselia2023learninguitocodereversegenerator} utilized advanced visual encoders and language decoders, fine-tuning the pipeline with visual similarity signals. However, their examples primarily included simple elements. \citet{si2024design2code, laurençon2024unlockingconversionwebscreenshots} firstly study whether VLMs can transform real-world screenshots to HTML webpages in an end2end pattern and demonstrate promising initial results. However, using screenshots as input is still unrealistic in the UI coding workflow. \citet{zhang2024frontend} shows one of the first demonstrations of leveraging VLMs in the sketch-to-code transformation without comprehensive benchmarking and framework design.

\section{The Sketch2Code Benchmark}

\subsection{Data Collection}
We curated our sketch dataset based on a diverse set of 484 real-world webpages collected by \citet{si2024design2code} under ODC-By license \footnote{We will release our dataset under the same
license, which is intended for research use only.}. Sketches are drawn following the standard wireframing conventions\footnote{\url{https://balsamiq.com/learn/articles/what-are-wireframes/}} by annotators recruited on Prolific\footnote{\url{www.prolific.com}}. Annotators are selected based on their self-reported expertise in UI design and their drawings of three sample sketches in a qualifier study. We selected 21 annotators from 723 total participants in the qualifier run. We then asked each selected annotator to draw 20-60 sketches of different webpages. Participants are compensated for \$2 during the qualifier and \$20/hr for the main study.

We have collected a total of 731 sketches for 484 webpage screenshots. To avoid overfitting to a particular style of sketches, we assigned a subset of the webpages to multiple annotators with varying styles and qualities. In particular, 18.0\% of the webpages are sketched by 2 designers, 16.5\% of the webpages are sketched by 3+ designers, and the remaining webpages are sketched by a single designer. Due to budget limits, we could not assign multiple designers to all webpages in the source dataset.
Appendix Figure \ref{fig:data_samples} contains example sketch-screenshot pairs of our dataset.

\begin{figure*}[t]
  \centering
  \includegraphics[width=\textwidth]{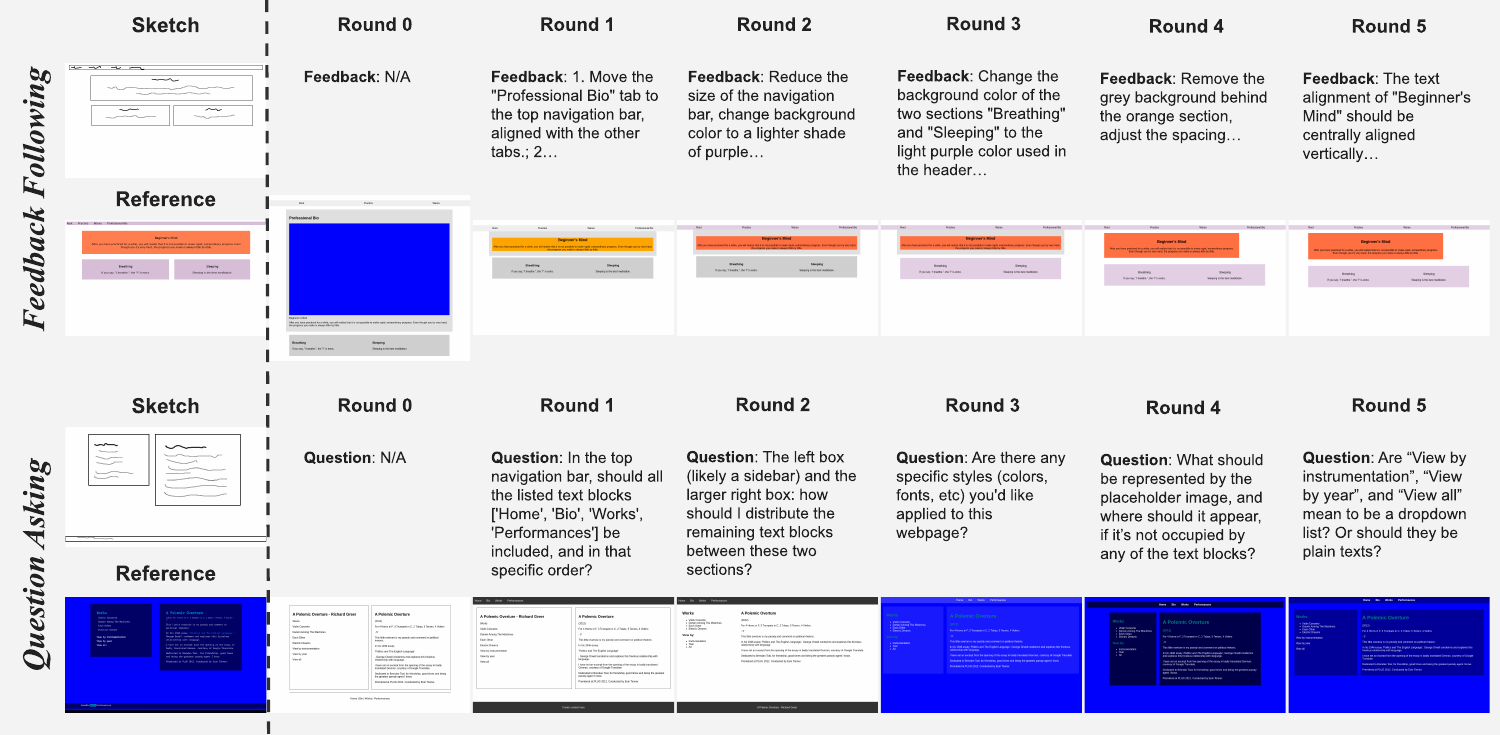}
  \caption{Multi-turn generation examples using GPT-4o, where we can observe the generated webpages get more similar to the reference as incorporating more feedback/answers.}
  \vspace{-1em}
  \label{fig:example_generations}
\end{figure*}

\subsection{Task Definitions}
\paragraph{Baseline: Direct Generation} In the simplest format, sketch2code agents are given only the sketch (or together with the text content) and are asked to generate an HTML implementation directly.
The agents are allowed to use placeholders if the text content is not given.

However, such a task has inherent limitations. Since sketches are low-fidelity abstractions of UI designs, it is often impossible for the agent, or even human experts, to perfectly implement a frontend UI from sketch alone in a single turn. To successfully understand and implement visual designs via sketches, the agent must communicate effectively with the user to clarify design requirements and improve their implementations. Therefore, in addition to direct generation, we present two multi-turn interactive evaluation tasks: \textbf{feedback following} and \textbf{question asking}. In both scenarios, we follow such a setup: a (simulated) user has an intended web UI design that they hope to implement (i.e., the reference webpage), but they only have a rough sketch as a starting point. The sketch2code agent will be required to effectively engage in multi-turn communication with the user to clarify the requirements and collectively figure out an implementation of the intended/desired webpage, while a sketch along with all texts that should appear on the webpage initially.  Figure \ref{fig:benchmark_examples} shows an example of each multi-turn evaluation scenario.

\paragraph{Feedback Following} At each round, the sketch2code agent is first asked to generate a webpage given the currently available information. An LLM-simulated user will then compare the screenshot of the current implementation against the screenshot of the reference webpage and provide feedback instructions on improving the implementation. The agent is expected to incorporate the feedback into its implementation in the next round of interaction. This task follows most of the existing conversation AI workflows. We use the most capable model with full design details (the sketch, the screenshot of the reference webpage, and the actual HTML code for the reference webpage) to build the simulated users; more details are available in Appendix \ref{appendix:lazy_user}.

\paragraph{Question Asking} At each round, the sketch2code agent is instructed to proactively look for ambiguities and uncertainties within the sketch design and ask clarifying questions. The LLM-simulated user will answer the questions, and the agent will generate a new webpage at each turn based on the user's answer. Unlike most existing conversational agents, in this task, the agent is required to take on more cognitive workload and proactively initiate conversations instead of passively waiting for user instructions. Human evaluations of the simulated user's capabilities are detailed in \S~\ref{sec:human_annotation}.

\begin{table*}[t]
\small
    \centering
    \begin{tabular}{ll|cccc|c}
        \toprule
        \textbf{Model} & \textbf{Prompting} & \textbf{Layout Sim.} & \textbf{Text IoU} & \textbf{Image IoU} & \textbf{Other IoU} & \textbf{Human Sat.} \\
        \midrule
        GPT-4o & Direct & 19.20 & 17.12 & \textbf{16.19} & \textbf{3.03} & 30.0 \\
         & Text-Augmented & 21.33 & 22.08 & \textbf{13.23} & 2.75 & - \\
        \midrule
        GPT-4o-Mini & Direct & 11.49 & 13.51 & 2.36 & 1.27 & 12.0 \\
         & Text-Augmented & 16.25 & 20.84 & 0.72 & 1.12 & - \\
        \midrule
        \midrule
        Claude-3-Opus & Direct & 12.86 & 10.43 & 12.67 & 0.65 & 10.0 \\
         & Text-Augmented & 17.11 & 18.09 & 8.32 & 2.97 & - \\
        \midrule
        Claude-3.5-Sonnet & Direct & \textbf{21.64} & \textbf{22.51*} & 10.47 & 2.94 & \textbf{36.0} \\
         & Text-Augmented & \textbf{22.26} & \textbf{25.33*} & 9.21 & \textbf{3.58} & - \\
        \midrule
        Claude-3-Sonnet & Direct & 11.97 & 10.61 & 10.09 & 0.73 & 0.0 \\
         & Text-Augmented & 14.22 & 15.85 & 6.62 & 1.72 & - \\
        \midrule
        Claude-3-Haiku & Direct & 10.25 & 12.61 & 3.15 & 1.17 & 6.0 \\
         & Text-Augmented & 17.52 & 20.60 & 2.72 & 2.22 & - \\
        \midrule
        \midrule
        Gemini-1.5-Pro & Direct & 18.25 & 16.44 & 14.69 & 1.12 & 22.0 \\
         & Text-Augmented & 18.72 & 19.46 & 11.79 & 0.96 & - \\
        \midrule
        Gemini-1.5-Flash & Direct & 14.15 & 13.28 & 8.77 & 0.03 & 8.0 \\
         & Text-Augmented & 15.22 & 13.25 & 7.81 & 0.16 & - \\
        \midrule
        \midrule
        InternVL2-8b & Direct & 10.08 & 11.28 & 6.13 & 0.00 & 2.0 \\
         & Text-Augmented & 4.01 & 4.89 & 1.41 & 0.60 & - \\
        \midrule
        Llava-1.6-8b & Direct & 6.68 & 6.91 & 3.43 & 0.36 & 0.0 \\
         & Text-Augmented & 8.00 & 9.26 & 1.95 & 0.57 & - \\
        \bottomrule
    \end{tabular}
    \vspace{-1em}
    \caption{The performance of eight commercial and two open-source models on the Sketch2Code direct generation task. The Layout Similarity is computed as the weighted average of the IoU for each visual component. The human satisfaction rate is the percentage of generation outputs labeled as "Satisfactory/Close Match" by human annotators. * indicates statistical significance (p<0.05) comparing to the second best performing model.
    }
    \label{tab:direct}
\end{table*}

\section{Experiments}
\label{sec:experiments}

\subsection{Prompting Methods}
% \paragraph{Direct Prompting}
We evaluated the agents using two prompting methods: \textbf{direct prompting} and \textbf{text-augmented prompting}. In direct prompting, the agent is provided with a sketch design only and is asked to generate an HTML prototype without access to additional information.
% \paragraph{Text-Augmented Prompting}
% Direct prompting assumes the agent has access to the only a single layout sketch. 
% In reality, however, the user often has some idea of what concrete content to put on the webpage in addition to the abstract layout. To simulate this behavior in our experiments, 
In text-augmented prompting, we augment the agent with all text content extracted from the reference webpage as part of the initial user prompt, following \citet{si2024design2code}. The exact prompts used and additional experiment details are available in Appendix \ref{appendix:experiment_details}.

Preliminary results show that text-augmented prompting is a more realistic setup and yields the most stable outputs across multiple interaction turns. Hence, we use text-augmented prompting as the starting point for all multi-turn experiments.

\subsection{Evaluation Metrics}
\label{sec: layout similarity}

\paragraph{Visual Similarity}
Following \citet{si2024design2code}, we calculate the visual similarity score as the average of Block Match, Text, Position, Color, and CLIP scores, which gives a complete assessment of the generated complete webpages.

However, compared to using screenshots, it is not meaningful to compare websites generated using sketches with reference websites since most textual and stylistic information is not provided.
To this end, we propose an IoU(Intersection over Union)-based metric that focuses solely on layout similarities.

\paragraph{Layout Similarity}
Given a generated and a reference HTML, we would first extract a list of visual components from each HTML file. To calculate the overlap within the same type of components, we identified seven classes of higher-level visual component types: \textit{text blocks, images, video containers, navigation bars, forms/tables}, and \textit{buttons}. Detailed explanations for each visual component type and their corresponding HTML tag selectors are available in Appendix \ref{appendix:layout_sim}.

For each visual component type $c$, we define its layout similarity as the IoU of the total area taken by all bounding boxes of components with type $c$ in the reference \& generated webpages:
$$IoU(c) = \frac{A'_c \cap A_c} {A'_c \cup A_c}$$

Where $A_c$ and $A'_c$ are the areas taken by components with type $c$ in the reference and generated webpages, respectively.

The overall layout similarity between two webpages is the weighted average of IoU scores of all visual component types $c \in \mathbf{C}$.
$$Sim_{Layout} = \sum_{c \in \mathbf{C}} {\frac{A'_{c} + A_{c}}{\sum_{c' \in \mathbf{C}} {(A'_{c'}+A_{c'})}} \times IoU(c)}$$

In practice, we evaluate direction generation only using the layout similarity metric. For other settings (Feedback-following and Question-asking), since agents have access to textual content initially and are available to collect more information through multi-turn conversation, we use both visual similarity and layout similarity metrics.

% \paragraph{Structural Grounding Prompting}
% Understanding and extracting visual layouts from sketches proves to be challenging. Preliminary experiments showed that even the strongest models struggle to extract consistent layout information from abstract sketches. To enhance layout understanding, we further include \textbf{structural grounding prompting} into our evaluation: ask an agent to first translate the sketch layout into a JSON data structure before generating the actual HTML code.
% Although the grounding improves the single-turn performance, we found that it doubles the length of the context and often leads to less consistent multi-turn outputs.
% Detailed JSON schema are available in Appendix \ref{appendix:layout_schema}.

% \ryan{Should I add another paragraph here elaborating how the prompts work under multi-turn interactions? Should I frame the multi-turn interactions in some RL terminologies?} Yanzhe: should be fine

\subsection{Result: Direct Generation}
\label{sec:results_direct}
We present direct generation results for all 10 evaluated models in Table \ref{tab:direct}. These include 8 commercial models (GPT-4o, GPT-4o mini, Gemini 1.5 Pro, Gemini 1.5 Flash, Claude 3.5 Sonnet, Claude 3 Opus/Sonnet/Haiku), and 2 open-source models (Llava-1.6-8b and InternVL2-8b). We choose not to run larger open-source models due to the high computational cost of generating hundreds of thousands of tokens per run.

First, we observed a considerable gap between commercial and open-source models on direct sketch-to-code generations. All commercial models outperformed open-source models, and the open-source models rarely achieves a layout similarity higher than \textbf{10\%}. While text-augmented prompting boosts the performance of all commercial models, the additional context can sometimes be detrimental to the generation qualities of open-source models. With direct prompting, Llava-1.6 and InternVL2 can generate correctly formatted HTML outputs \textbf{82.4\%}/\textbf{98.0\%} of the time. However, after giving the textual content on the webpage, they tend to generate repetitive content without finishing and can only output the correct HTML format \textbf{66.7\%}/\textbf{39.2\%} of the time, respectively.

Among the eight commercial models, Claude 3.5 Sonnet and GPT-4o lead the performance in single-turn generations, achieving the best layout similarity with direct and text-augmented prompting. While text-augmented prompting can \textbf{significantly} boost their performance by \textbf{5-7\%} for smaller, less capable models (e.g., GPT-4o Mini and Claude 3 Haiku), such effect dwindles with more capable models, suggesting that strong models can generate correct layouts without referencing text content.

In addition to the automated metrics, we conducted a human evaluation with Prolific crowd annotators for the direct prompting outputs. For each generated result, the annotators are given three options, indicating whether the generated layout is "Satisfactory/Close Match", "Loosely Match with Minor Fixes", or "Unsatisfactory". The human satisfaction rate had a $r^2$ value of \textbf{0.87} (p=0.00008), and a Kendall's Tau score of \textbf{0.72} (p=0.004) with the IoU-based layout similarity metric, indicating a strong correlation between the automated metric and human judgment. More details in Appendix~\ref{appendix:layout_sim}.

\begin{figure*}[ht]
\centering
\includegraphics[width=\textwidth]{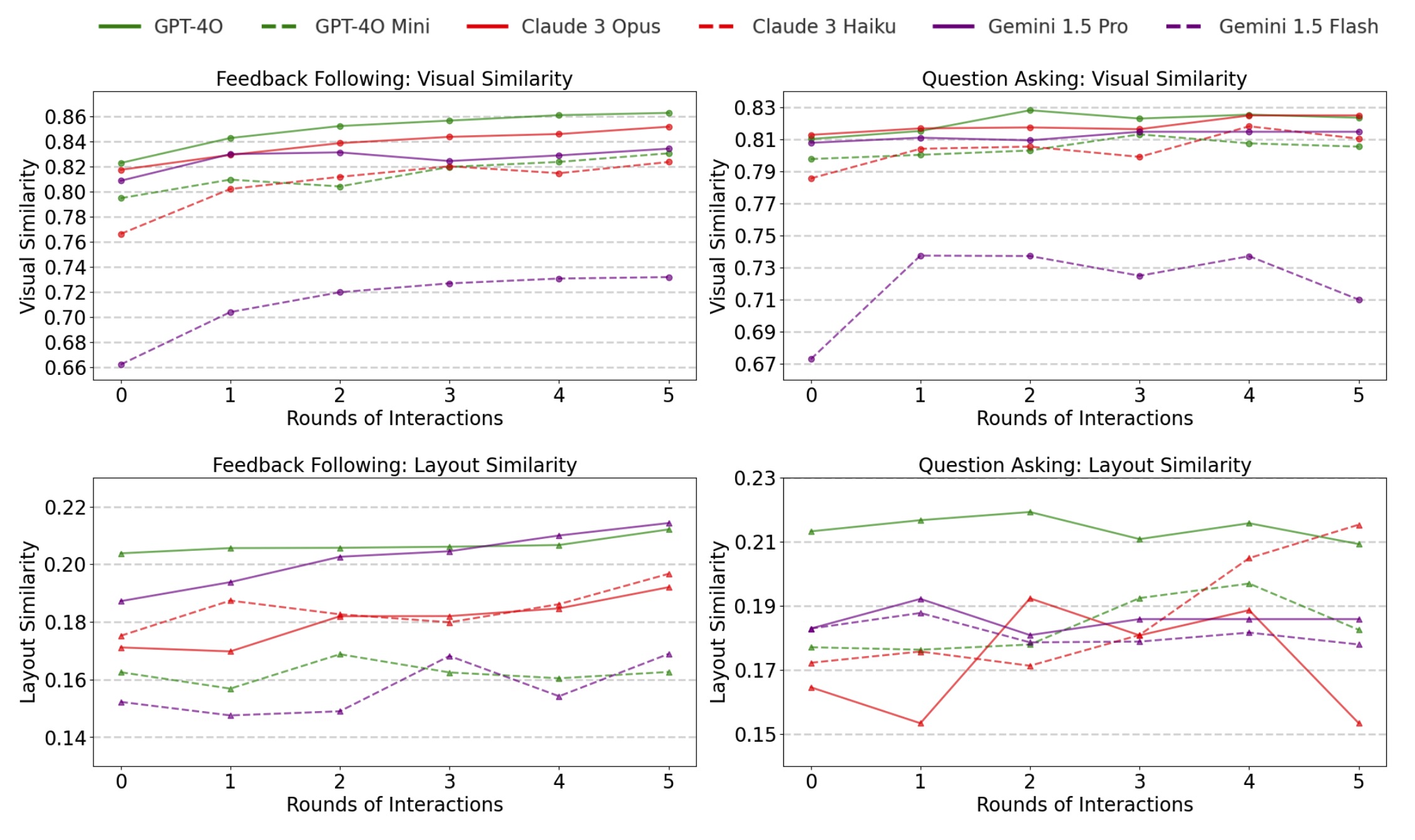}
\caption{The performances of six models on the feedback following benchmark (left) and the question asking benchmark (right): GPT-4o, GPT-4o Mini, Claude-3-Opus, Claude-3-Haiku, Gemini 1.5 Pro, and Gemini 1.5 Flash.}
\vspace{-1em}
\label{fig:performance_overview}
\end{figure*}

\subsection{Result: Multi-turn Evaluation}
\label{sec:results_multiturn}
Figure \ref{fig:example_generations} shows the example outputs of GPT-4o on the two multi-turn evaluation tasks. We present the performance of the largest and smallest models in each of the three commercial model families (GPT-4o, Gemini 1.5, and Claude 3) in Figure~\ref{fig:performance_overview}. \textbf{(I) All models displayed noticeable improvements in feedback following}. The best commercial models achieves improvements of up to \textbf{7.1\%} in visual similarity and \textbf{2.7\%} in IoU-based layout similarity within five rounds of interaction. Models with weaker single-turn performance can sometimes lead to more multi-turn improvements. \textbf{(II) Question asking is more challenging} as all models struggled to pose effective questions about the sketches and showed very few improvements with statistical significance.
% In particular, Gemini 1.5 Pro was rather reluctant to ask questions, often prematurely stopping before reaching the maximum five-question limit. Table \ref{tab:qa} shows the performance breakdowns for each model in question asking. 
Most performance gains occurred within the first two rounds of interactions, and performance often plateaued or even deteriorated after three to four rounds of interactions.
We provide a more detailed analysis of the types of effective versus ineffective questions asked in \S~\ref{sec:question-analysis}.

\noindent
\paragraph{Case Study} Figure \ref{fig:performance_claude} further shows the performance of the four models in the Claude model family. (I) While we observe an apparent scaling-up effect within the Claude 3 models regarding visual similarity (Haiku $<$ Sonnet $<$ Opus), such effect is not observable in layout similarity as Sonnet and Opus sometimes perform worse than Haiku. Since the visual similarity is heavily based on text content matching while layout similarity only considers spatial overlap, we assume \emph{solely scaling up model sizes barely helps with the layout understanding and generation compared to text-related capabilities}. (II) While Claude 3.5 Sonnet excels at generating webpage layout (layout similarity score above 22\%), it struggles to leverage information from user interactions. We assume it might be over-optimized for single-turn generations. Additional details on direct \& multi-turn evaluation results are available in Appendix~\ref{appendix:additional_experiment_results}.

% According to the scoring breakdown in Table \ref{tab:user-feedback}, models tend to show more significant and consistent improvements in text layout (Position, Text IoU) than in image layout (Image IoU). This is likely \yanzhe{why possibly? do we have numbers?} \ryan{This is mostly from my qualitative analysis on a subset of the examples. I could run a GPT classification of text feedback v.s. image feedback, but that would also seem ad hoc?} due to the simulated user prioritizing text layout and formatting over the layout of images in its provided feedback.

\subsection{Human Annotation}
\label{sec:human_annotation}
To ensure the validity of our automated metrics and the simulated user environment, we conducted a series of human evaluations on different components across our evaluation pipeline. Details on the recruitment process are available in Appendix~\ref{appendix:human_annotation}.

\noindent
\paragraph{Layout Similarity} We leveraged pairwise comparisons \citep{zhou2023lima, dubois2024alpacafarm} to evaluate how well the automated layout similarity metrics align with human judgement and preferences. Each annotator is given pairs of generated implementations of the same reference webpage and is asked to select the one that is more similar in layout to the reference. 
Our IoU-based layout similarity metrics agreed with human judgment 69.2\% of the time, comparable to the agreement score between human annotators, 65.8\%. The five annotators achieved a Fleiss Kappa score of 0.47, a relatively high agreement for subjective tasks \cite{landis1977measurement}.
% We would like to point out that comparing layout similarities and differences is sometimes a rather challenging and subjective task for human annotators, especially with webpages varying in color schemes or aspect ratios.

\noindent
\paragraph{Simulated QA} To verify the answer quality of the simulated user during the question-asking benchmark, participants are presented with a sketch design and a reference webpage, followed by a single round of simulated QA.
According to the human annotators, 93.3\% of the generated answers directly respond to the given question, and 86.7\% of the answers remain faithful to the reference webpage.
Another validity check is whether the generated answer contains any code leakage or direct references to HTML elements. We found that the simulated user exhibits code leakage issues in 8.3\% of the answers. Fleiss Kappa scores and additional details are available in Appendix~\ref{appendix:human_annotation}.

\noindent
\paragraph{Simulated User Feedback}
% We leveraged direct human assessment again to evaluate the quality of the simulated user feedback in the feedback-following scenario. 
Participants are given a current implementation from a model, a reference webpage, and feedback generated by the simulated user and are asked to rate the simulated feedback.
Our annotators found that 86.7\% of the simulated feedback is easy to follow, and 83.3\% of the simulated feedback accurately points out the difference between the current \& reference implementations.

\section{User Study and Analysis}
\label{sec:user_study}
To better understand the importance of sketching in the UI/UX development cycle, and the potential use cases and implications of a sketch2code agent, we conducted a user study by interviewing eight UI/UX experts recruited from Upwork\footnote{\url{https://www.upwork.com/}}.

All experts agreed that low-fidelity sketches play a substantial role in modern UI/UX development. Furthermore, they found that a sketch2code agent would significantly benefit their work. A sketch2code agent can help users quickly flesh out early-stage ideas and break the communication barrier between clients and designers. More interestingly, seven out of eight participants showed \textbf{strong preference} towards the question-asking agent. The interviewees expressed that they needed to specify every design detail to an agent that passively follows user instructions. In contrast, a question-asking agent can take over most of the cognitive workload, and the user only needs to focus on the parts being asked. Three of the experts pointed out that \textbf{the agent can proactively “guide” the user through certain design decisions \& choices via a series of targeted questions}, so the user do not have to figure out every single detail themselves. Moreover, the participants mentioned that it is difficult to select visual components and specify visual information with natural language feedback/answers alone. When communicating design ideas in real life, people can simply point to specific visual components by mouse or by finger instead of using words, which is faster, easier, and more reliable. Detailed findings are available in Appendix~\ref{appendix:user_study}.

% \paragraph{O4: Additional desired features for the sketch2code agent.}
% \ryan{Should we move this to future work?}
% In addition to enhancing the models' capabilities, the participants have pointed out other desirable features that should be included in the sketch2code agent. All five participants mentioned that they wish there are simpler and more controllable ways to edit certain visual components (e.g., quick selections for colors, fonts, button types) instead of having to specify everything in natural language. Three out of the five experts conveyed that the application should support state-sharing and online collaboration between multiple users/co-workers. And two of the interviewees hoped that they can export the generations to design softwares like Figma instead of having an HTML file.

% \section{Analysis}
% \vspace{-1em}

% \subsection{How to ask effective questions?}
% \label{sec:question-analysis}

%In \S~\ref{sec:experiments} and \S~\ref{sec:user_study}, we discussed that although the question-asking framework is a much desired interaction paradigm for human users, current SoTA VLMs struggles to find effective question. 

Moreover, we invited 5 UI/UX experts who participated in the user studies to (1) provide feedback on 100 generated designs from three models (GPT-4o, Claude-3-Opus, and Gemini-1.5-Pro) and (2) ask questions on 100 sketches. We further compare the difference between AI-generated and human-written feedback/questions below \footnote{We will release the collected expert annotations to facilitate future research.}.

\textbf{Simulated user feedback achieved comparable performance as feedback from real human experts}: On average, each feedback from the simulated user improves the agents' performance in visual similarity by 1.41\% and layout similarity by 1.05\%, whereas each feedback from a human expert improves the visual similarity by 1.62\% and layout similarity by 0.98\%.
%Similar to \S~\ref{sec:question-analysis}, we summarized the feedback \& instructions provided by the simulated user into seven broad categories. Detailed results and taxonomy are available in Appendix \ref{appendix:question_classification}.
We developed a taxonomy and summarized the user feedback into seven categories (Detailed taxonomy and statistics are available in Appendix \ref{appendix:question_classification}.).
Interestingly, human experts provide more feedback on colors and styling, while simulated users focus more on texts and general comments about the webpage.
Furthermore, we evaluated the effectiveness of each type of simulated feedback in Appendix Table \ref{tab:feedback_improv_by_type}, where we explicitly prompt the simulated user to give different types of feedback. We found that feedback on layout structure is most helpful for all three models, with the most significant average improvements in layout and overall visual similarity. This is followed by feedback on the styling and layout of major visual components, where all three models can again show positive average improvement in both visual and layout scores.
% \vspace{-1em}
% While instructions on text formatting do not often facilitate layout improvements, they improve the visual scores for GPT-4o and Claude-3-Opus. Models do not usually show improvements in either layout or visual scores from feedback on tertiary elements.

\label{sec:question-analysis}
\textbf{The effectiveness of questions asked by models lags far behind that of experts.}
% \ryan{Do I need to explain that these are the same people in the user studies? What other information should I provide here?} We ran these examples 
Initially, the models' performance on average \textbf{decreases} by 1.12\% in visual similarity and improves by 0.74\% in layout similarity after asking each question. However, by replacing the agent-generated questions with questions written by human experts, the models were able to \textbf{improve} their visual similarity by 0.58\% and layout similarity by 1.49\% with each question.
Based on a taxonomy of the nine most common types of questions (The full taxonomy and further details in Appendix \ref{appendix:question_classification}.), we found that human experts often ask about the general styling and layout of the webpage, whereas the agents often focus on the specific placements of textual contents or ask irrelevant questions.
As shown in Appendix Table \ref{tab:feedback_improv_by_type}, the most effective type of questions are questions regarding stylistic choices of primary visual components, followed by questions regarding the general layout or the positional placements of major elements. Appendix~\ref{appendix:question_classification} contains a more detailed analysis of the effectiveness of different questions.

\section{Conclusion and Future Work}
\label{sec:conclusion}
In this work, we introduced Sketch2Code, a novel interactive evaluation framework that assesses Vision Language Models' (VLMs) capability for multi-turn front-end UI/UX automation. We proposed two interaction paradigms: \textbf{feedback following} and \textbf{question asking}. Our evaluations revealed that while modern commercial VLMs perform reasonably well in feedback following—improving visual and layout similarities by 3\% and 1.8\% over five interaction rounds—they struggle with asking meaningful questions. However, a user study with eight UI/UX practitioners showed a user preference for the question-asking paradigm, as it allows the agent to take on more cognitive responsibilities, highlighting a gap between user expectations and current model capabilities.

We outline future research directions inspired by Sketch2Code: (i) \textbf{Training open-source models} for multi-turn UI generations. This can be challenging due to the long contexts and multi-modality. To facilitate large-scale training, we present an automated pipeline for generating realistic synthetic sketches at scale (see Appendix~\ref{appendix:synthetic_sketches}). (ii) \textbf{Developing agentic frameworks} that are more capable of cognitive reasoning and proactively guiding users through multi-turn design workflows, instead of passively following instructions. (iii) \textbf{Creating end-to-end UI/UX AI applications} to enhance designer productivity and make UI designs more accessible to non-experts.

\section{Limitations}
Despite our efforts, we address the following limitations of our work:

\begin{itemize}
    \item Due to computational limitations, we evaluated only 8b open-source models, InternVL2-8b and Llava-1.6-8B. Larger open-source models were not included in this study due to the computational cost, which might have better multi-turn interaction capability.
    \item Second, the multi-turn evaluation pipeline with simulated users is computationally expensive. Running a single example in the feedback following/question-asking benchmarks requires 40,000 to 160,000 input tokens and approximately 10,000 output tokens, making large-scale evaluations costly, though manageable for specific use cases.
    \item Moreover, while the sketch2code agent converts natural language inputs to HTML code, user studies with UI/UX practitioners indicate a preference for more direct, deterministic ways (e.g., mouse clicks, drags) to select and modify visual components, as well as support for exporting outputs to Figma or other design software, highlighting the need for additional input/output modalities.
    \item Finally, we acknowledge the potential for misuse of this technology by malicious actors, who might generate harmful webpages or attempt to reverse-engineer code from proprietary or licensed websites.
\end{itemize}

% \section{Ethical Considerations}
% Something?

\section*{Acknowledgments}
We thank Yutong Zhang, Chenglei Si, Lin Qiu, Harshit Joshi, Yicheng Fu, John Yang, Ryan Louie, Yijia Shao, Aryaman Arora, Michelle Lam, Dora Zhao, Hao Zhu, Michael Ryan, Raj Shah, Will Held and other outstanding members in the SALT lab/Stanford NLP/Stanford HCI for their valuable feedback on different stages of this work.

\bibliography{custom}

% \newpage
\appendix

\begin{figure}[ht]
\centering
\includegraphics[width=0.35\textwidth]{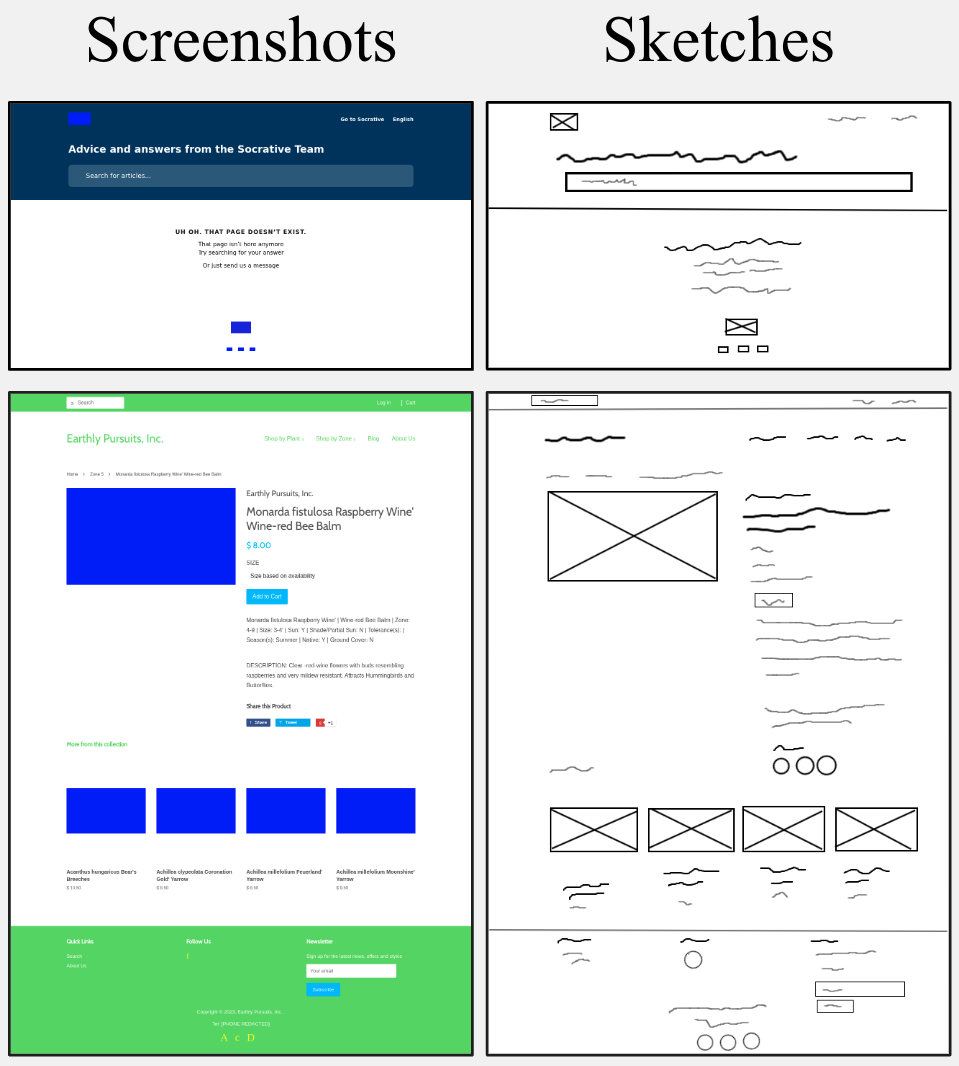}
\caption{Examples of screenshots (left) and human-drawn sketches (right) of the Sketch2Code dataset. Sketches are drawn following the wireframing conventions, where boxes with an "X" inside replace images, and curly lines represent texts.}
% \vspace{-1em}
\label{fig:data_samples}
\end{figure}

\begin{table*}[!ht]
\centering
\begin{tabularx}{\textwidth}{@{}l|X@{}}
\toprule
\textbf{Component Type} & \textbf{CSS Selector} \\ 
\midrule
Video          & \texttt{video} \\
\midrule
Image          & \texttt{img} \\
\midrule
Text Block     & \texttt{p, span, a, strong, h1, h2, h3, h4, h5, h6, li, th, td, label, code, pre, div} \\
\midrule
Form/Table     & \texttt{form, table, div.form} \\
\midrule
Button         & \texttt{button, input[type="button"], input[type="submit"], [role="button"]} \\
\midrule
Navigation Bar & \texttt{nav, [role="navigation"], .navbar, [class~="nav"], [class~="navigation"], [class~="menu"], [class~="navbar"], [id="menu"], [id="nav"], [id="navigation"], [id="navbar"]} \\
\midrule
Divider        & \texttt{hr, [class*="separator"], [class*="divider"], [id="separator"], [id="divider"], [role="separator"]} \\
\bottomrule
\end{tabularx}
\caption{HTML and CSS Selectors for Visual Components}
\label{tab:selectors}
\end{table*}

\begin{figure*}
\begin{lstlisting}[caption={Instructions given to Prolific participants for human evaluation on direct generation outputs}, label={lst:human_eval_instructions}, captionpos=b]
Task Overview
In this survey, you will be given a reference webpage's screenshot, as well as a model generated webpage that try to replicate the reference webpage's layout. Your task is to judge whether the generated webpage matches the layout of the reference webpage. The borders for each webpage is marked in black.

Note: All images in the original webpages are replaced by blue rectangles as placeholders. Often times, the generated webpages will use "lorem ipsum..." as placeholder texts. Please disregard these placeholders and treat them as normal texts instead.

Comparison Guide
In this survey, you should base your comparisons solely on the layout similarities between the example webpages and the reference webpage.

You should pay close attention to:
Visual component matching: a good generated example should contain all visual components that are present in the reference image and have no extra components. You should examine if the visual components in the candidate example match with the ones in the reference image. Visual components include text blocks, images, menus/navigation bars, tables, form inputs, etc.
Overall layout arrangement: a good generated example should have its visual components arranged similarly as the reference image. You should examine how much the layout arrangement in each candidate overlap with the reference layout.
Size and position of each visual component: you should pay attention to the relative size and position (w.r.t the webpage's width and height) of each visual component w.r.t. the overall width and height of the webpage. Sometimes, the webpages have different width/height ratios, but their main visual components have similar relative sizes and positions. When comparing text blocks, it is important that you consider only the sizes and positions of the text blocks. The actual text content should be disregarded.
You should ignore/not pay attention to:
Detailed text content: since the survey focuses solely on layout similarities, the exact details of textual content does not matter and should be factored out from your judgement. You should never rate a candidate example down if it contains placeholder texts, as long as text blocks are placed similarly as the reference image.
Color and Styles: background color, the color of each section/text block, the font/size/color of the text, or any other color & styling decisions should never impact your final choice. Your decisions should be based on solely the layout and placement of each component.

When making your judgement, you have three options:
1. **Satisfactory/Close Match**: The layout of the generated and reference webpages matches closely, with only minor differences in details. Major visual components (such as headers, images, text sections) align closely, and any variations are insignificant. The two pages can serve similar functions.

2. **Loosely Match with Minor Fixes**: There are some observable differences between the generated and reference webpages, but the overall layout of major visual components roughly matches. The observed differences between the webpage layouts can be addressed with minor fixes.

3. **Unsatisfactory**: There are significant differences or mismatches in the layout between the reference and generated webpages. There are major components misaligned, missing, or structured differently, resulting in a layout that does not closely resemble the reference. Significant changes are required to make the layouts match.
\end{lstlisting}
\end{figure*}

\begin{figure*}[htbp]
  \centering
  \includegraphics[width=0.85\textwidth]{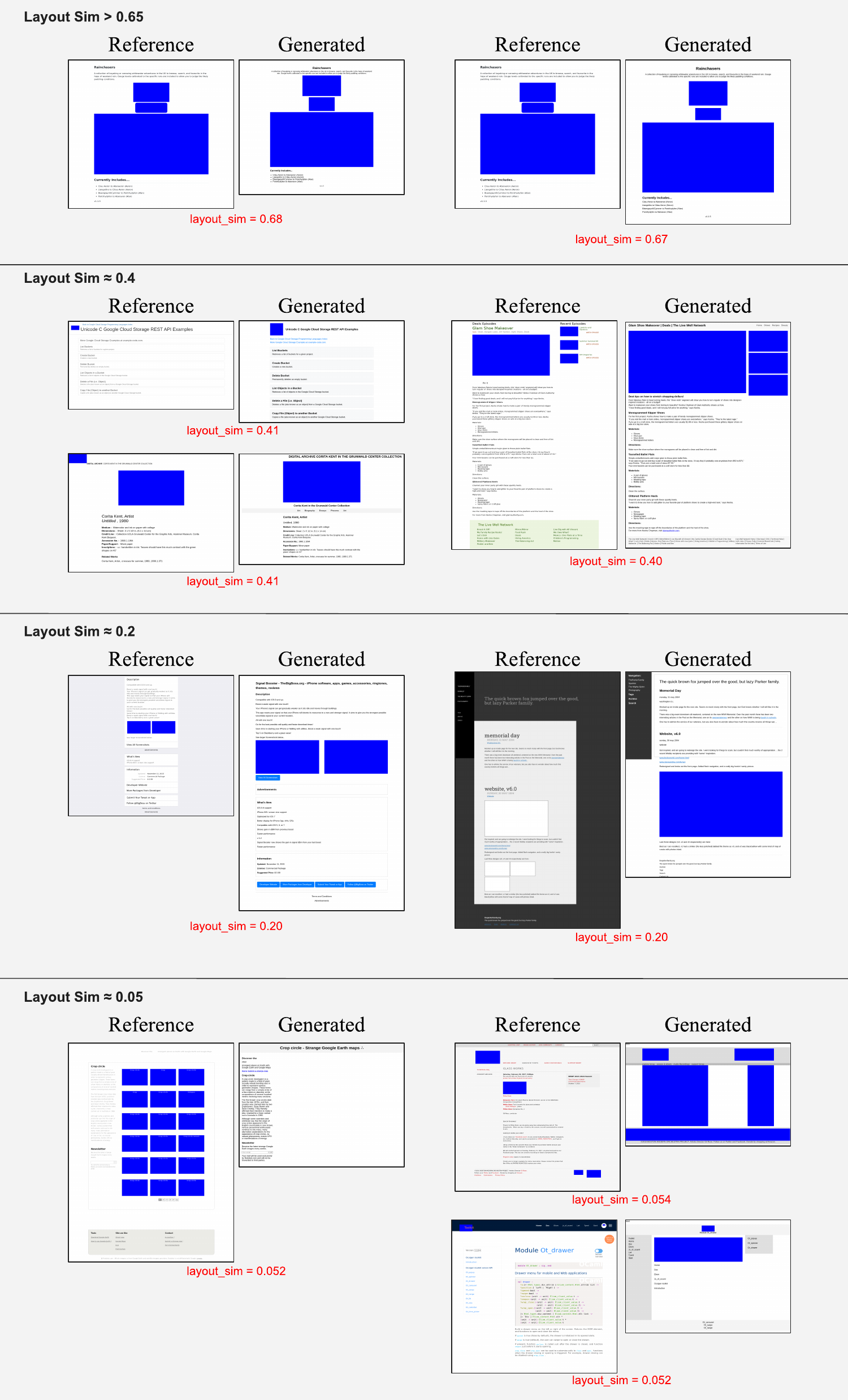}
  \caption{Example reference-generation pairs with different levels of layout similarity scores.}
  \label{fig:layout_sim_examples}
\end{figure*}

\begin{figure*}[ht]
\centering
\includegraphics[width=\textwidth]{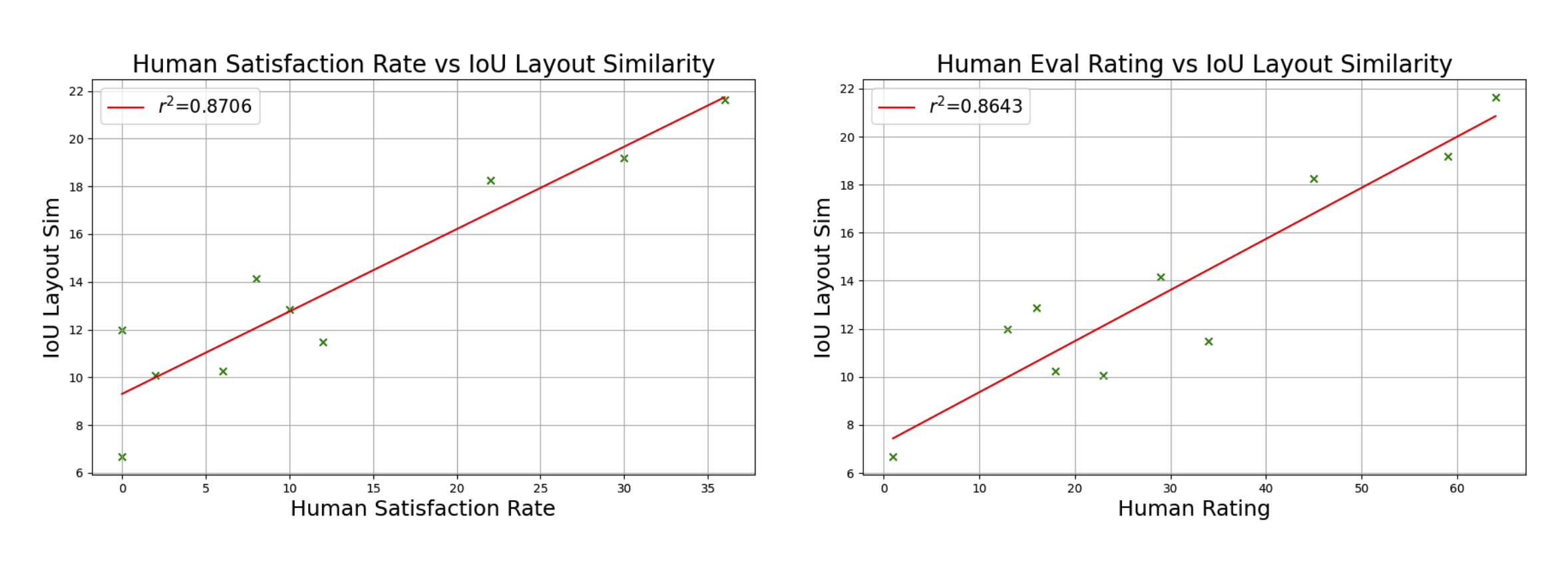}
\caption{Human satisfaction rate v.s. avg layout similarity (left) and aggregated human rating v.s. avg layout similarity (right) for the ten evaluated models.}
\label{fig:human_eval_corr}
\end{figure*}

\begin{table*}[!ht]
    \centering
    \begin{tabular}{l|ccc}
        \toprule
        \textbf{Model} & \textbf{Close Match} & \textbf{Loose Match} & \textbf{Unsatisfactory} \\
        \midrule
        GPT-4O & 30.0 & 58.0 & 12.0 \\
        GPT-4O Mini & 12.0 & 44.0 & 44.0 \\
        \midrule
        Claude 3 Opus & 10.0 & 12.0 & 78.0 \\
        Claude 3.5 Sonnet & 36.0 & 56.0 & 8.0 \\
        Claude 3 Sonnet & 0.0 & 26.0 & 73.0 \\
        Claude 3 Haiku & 6.0 & 24.0 & 70.0 \\
        \midrule
        Gemini 1.5 Pro & 22.0 & 46.0 & 32.0 \\
        Gemini 1.5 Flash & 8.0 & 42.0 & 42.0 \\
        \midrule
        InternVL2-8b & 2.0 & 42.0 & 56.0 \\
        Llava-1.6-8b & 0.0 & 2.0 & 98.0 \\
        \bottomrule
    \end{tabular}
    \caption{Human evaluation breakdown of the ten evaluated models under direct prompting. It shows the percentages of outputs labeled as "Satisfactory/Close Match", "Loosely Match with Minor Fixes", and "Unsatisfactory" per model by human annotators.}
    \label{tab:human_eval_breakdown}
\end{table*}

\section{Layout Similarity Details}
\label{appendix:layout_sim}
In order to compute visual similarities, we identified several higher-level visual component types and extracted a list of common HTML and CSS tags that are commonly used to represent each type of visual component. Table \ref{tab:selectors} shows a list of HTML \& CSS selectors for each type of visual component. Figure~\ref{fig:layout_sim_examples} presents example generations under different levels of layout similarity scores.

\paragraph{Human Evaluation} to further verify the reliability of our automated metric, we conducted human evaluations on the generated layout qualities for the ten evaluated model under direct prompting. We subsampled 50 direct generation outputs from each model for human evaluation. Prolific crowd annotators are screened based on the following criteria: 1. 2500+ completed studies; 2. 99\%+ acceptance rate; 3. fluency in English; 4. having a Bachelor or higher education degree. The participants are given pairs of reference v.s. generated webpages and are asked to decide if the generated webpage layout is "Satisfactory/Close Match", "Loosely Match with Minor Fixes", or "Unsatisfactory". Detailed definitions for each options as well as three examples are provided to the participants to calibrate the results. The detailed instructions are available in Listing~\ref{lst:human_eval_instructions}. Each output example is annotated by three participants, and we use majority votes to determine the final label for each generation output.

The percentage of "Satisfactory" generated outputs per model linearly correlates with the average layout similarity scores with $r^2$ value of 0.87 (p=0.0008) and Kendall's Tau score of 0.72 (p=0.004). In addition, we computed the average human \textbf{rating} per model by counting each "Satisfactory/Close Match" as 1 point, "Loosely Match with Minor Fixes" as 0.5 point, and "Unsatisfactory" as 0 point. The averaged human ratings achieve a $r^2$ score of 0.86 (p=0.00009) and Kendall's Tau score of 0.64 (p=0.009) with the IoU-based layout similarity metric. Figure~\ref{fig:human_eval_corr} shows the linear correlations of human satisfaction rate v.s. layout similarity and human rating v.s. layout similarity. Table~\ref{tab:human_eval_breakdown} presents the breakdown percentages of each human evaluation label per model.

\begin{figure*}
\begin{lstlisting}[caption={User prompt for simulated user in the question asking task}, label={lst:qa_user}, captionpos=b]
You have access to two images. One is a sketch layout of a webpage drawn in the wireframing conventions, and the other one is a screenshot of a reference implementation. Please note that some images have already been replaced by placeholders (i.e., "rick.jpg") in the screenshot.

In addition, you also have access to the HTML implementation of the reference webpage:
```
{HTML_CODE}
```

-------------------------------------------------
Now, please answer the agent's questions based on the information you have. The agent will ask questions about elements in the sketch, and your answers **MUST** be **strictly** based on the provided images and html code.

Remember, you must answer the questions accurately and succinctly. You should **NEVER** make things up or provide any information more than what the agent asks for. The agent is not supposed to know about the reference implementation or its screenshot, so you should **NEVER** mention the reference implementation or the screenshot in your response, nor should you ever give out any HTML content to the agent. For example, if the agent asks for an element, you should answer with what is visible on the rendered webpage instead of the actual HTML tag or id. If the user asks for the color of something, you should describe the color in natural language (e.g., blue) instead of the hexadecimal color code. And if the user asks for the specific texts within a text block or paragraph, you should respond with a concise summary of the paragraph instead of reciting the text verbatim. You may acknowledge the fact that `rick.jpg` is used as image placeholders.

Format your answer to each question as a single sentence without omitting important information.

Agent Question:
\end{lstlisting}
\end{figure*}

\begin{figure*}
\begin{lstlisting}[caption={User prompt for simulated user in the user feedback following task}, label={lst:uf_user}, captionpos=b]
Suppose you are a frontend designer working with a code agent to implement an HTML webpage. You are provided with two images: the first image is the webpage you are hoping to produce, and the second one is the current implementation from the code agent. Note that images have already been replaced with blue rectangles as the placeholder.

Your job is to carefully compare the code agent's implementation against the intended webpage, and provide feedback to help the code agent make its implementation closer to the indended webpage. Your feedback should be specific to the differences in layouts and visual components on the two webpages. Please note that the code agent **DOES NOT** have access to the intended webpage, so you make sure to describe the intended visual components and where exactly the agent got wrong, instead of saying something like "refer to the format of the intended webpage". You should prioritize making sure that the code agent understands the correct layout before giving out any styling advice.

Limit your feedback to a single sentence.

You may compare and analyze the two webpages step by step. Once you are ready, your final feedback using triple quotes:
Feedback: """
{{YOUR_INSTRUCTIONS_HERE}}
"""

If you think the current implementation is close enough to the intended webpage, please output "Generation Complete" as your feedback. I.e.,
Feedback: """
Generation Complete
\end{lstlisting}
\end{figure*}

\section{LLM-Simulated user}
\label{appendix:lazy_user}

To automatically simulate and evaluate multi-turn conversations with sketch2code agents, we deployed a LLM-simulated user for each of our two multi-turn evaluation tasks. In both tasks, the simulated user is backed by a GPT-4o model with temperature=0. In the question asking task, the simulated user is given a sketch, a screenshot, and the HTML code for the screenshot, in order to provide the most accurate information to each question. In the feedback-following task, however, the simulated user is given only a sketch and a screenshot, as it is supposed to base its feedback solely on visual comparisons. "Laziness" is achieved by prompting the agent to answer all questions with at most one succinct sentence. Listing\ref{lst:qa_user} shows the user prompt for the simulated user in question asking, and Listing\ref{lst:uf_user} shows the user prompt for the simulated user in the feedback-following task.

\section{Experiment Details}
\label{appendix:experiment_details}

We used the same hyperparameter settings for all eight commercial models, with temperature = 0.0, max tokens = 4096, top p = 1.0, frequency/repetition penalty = 0.0, and presence penalty = 0.0.

For the two open-source model, we found that the models tend to self-repeat and output invalid HTML format. To improve the generation success rate, we adjusted the repetition penalty to 1.1, and temperature to 0.5 (while keeping other hyperparameters the same), and used best-of-3 sampling for all experiments.

Listing~\ref{lst:direct_generation} shows the system and user prompts used for direct generation tasks. Listing~\ref{lst:text_augmented} shows the system and user prompts used for text-augmented prompting.

For the feedback following multi-turn evaluation framework, we used the same text-augmented prompting to generate the first implementation prototype. Then, in each round of interaction, we will prompt the simulated user to provide a feedback for the current prototype, and then append the feedback as a user message to the sketch2code agent's conversation history to generate the next prototype.

For the question-asking evaluation framework, we first prompt the sketch2code agent to generate one or more questions about the given sketch. After the simulated user has answered each question, we will augment the agent with the question-answer pair and prompt it to generate a new implementation prototype. Listing~\ref{lst:question_generation_prompt} and Listing~\ref{lst:qa_augmented_prompt} shows the detailed agent prompts used for the question-asking benchmark.

Conducting multi-turn evaluations is a computationally expensive task. For example, it takes around 10 minutes and 50,000 tokens to run the full feedback following pipeline on a single data sample with GPT-4o. Given such constraint, we limited our multi-turn experiments to a randomly selected subset of 50 data samples from the 731 total sketches in the Sketch2Code dataset.

\begin{figure*}
\begin{lstlisting}[caption={Direct generation prompt for sketch2code agents, the topic embedded in user prompt is extracted from the HTML page title.}, label={lst:direct_generation}, captionpos=b]
System Prompt: You are an expert web developer who specializes in HTML and CSS. A user will provide you with a sketch design of the webpage following the wireframing conventions, where images are represented as boxes with an "X" inside, and texts are replaced with curly lines. You need to return a single html file that uses HTML and CSS to produce a webpage that strictly follows the sketch layout. Include all CSS code in the HTML file itself. If it involves any images, use "rick.jpg" as the placeholder name. You should try your best to figure out what text should be placed in each text block. In you are unsure, you may use "lorem ipsum..." as the placeholder text. However, you must make sure that the positions and sizes of these placeholder text blocks matches those on the provided sketch.

Do your best to reason out what each element in the sketch represents and write a HTML file with embedded CSS that implements the design. Do not hallucinate any dependencies to external files. Pay attention to things like size and position of all the elements, as well as the overall layout. You may assume that the page is static and ignore any user interactivity.
-------------------------------------------------

User Prompt: Here is a sketch design of a webpage about {topic}. Could you write a HTML+CSS code of this webpage for me?

Please format your code as
```
{{HTML_CSS_CODE}}
```
Remember to use "rick.jpg" as the placeholder for any images
\end{lstlisting}
\end{figure*}

\begin{figure*}
\begin{lstlisting}[caption={Text-augmented prompting for sketch2code agents.}, label={lst:text_augmented}, captionpos=b]
System Prompt: You are an expert web developer who specializes in HTML and CSS. A user will provide you with a sketch design of the webpage following the wireframing conventions, where images are represented as boxes with an "X" inside, and texts are replaced with curly lines. You need to return a single html file that uses HTML and CSS to produce a webpage that strictly follows the sketch layout. Include all CSS code in the HTML file itself. If it involves any images, use "rick.jpg" as the placeholder name. You should try your best to figure out what text should be placed in each text block. In you are unsure, you may use "lorem ipsum..." as the placeholder text. However, you must make sure that the positions and sizes of these placeholder text blocks matches those on the provided sketch.

Do your best to reason out what each element in the sketch represents and write a HTML file with embedded CSS that implements the design. Do not hallucinate any dependencies to external files. Pay attention to things like size and position of all the elements, as well as the overall layout. You may assume that the page is static and ignore any user interactivity.
-------------------------------------------------

User Prompt: Here is a sketch design of a webpage drawn in the wireframing conventions. In addition, here is a list of text blocks that I would like to include in the webpage:

{texts}

Could you write a HTML+CSS code of this webpage for me?

Please format your code as
```
{{HTML_CSS_CODE}}
```
Remember to use "rick.jpg" as the placeholder for any images
\end{lstlisting}
\end{figure*}

\begin{figure*}
\begin{lstlisting}[caption={System and user prompts used for question generation in the question-asking evaluation benchmark.}, label={lst:question_generation_prompt}, captionpos=b]
System Prompt: You are an expert web developer who specializes in HTML and CSS. A user will provide you with a sketch design of the webpage drawn in the wireframing conventions, where images are replaced by boxes with an "X" inside and texts are represented by curly lines. You need to return a single html file that uses HTML and CSS to produce a webpage that strictly follows the sketch design. Include all CSS code in the HTML file itself. If it involves any images, use "rick.jpg" as the placeholder. Some texts are replaced by curly lines as placeholders. You should try your best to infer what these texts should be, but do not hallucinate if you are not sure.

If you are unsure what certain elements are in the provided sketch, you should ask the user to clarify. Once you are confident, output a single HTML file with embedded CSS. Do not hallucinate any dependencies to external files. Pay attention to things like size and position of all the elements, as well as the overall layout.
-------------------------------------------------

User Prompt: Here is a sketch design of a webpage drawn in the wireframing conventions. In addition, here is a list of text blocks that I would like to include in the webpage:

{texts}

Could you write a HTML+CSS code of this webpage for me?

Remember, If you are uncertain about something, please ask clarification questions. Your questions should be thoughtful and specific, and you should ask no more than five questions in each turn.

If you want to ask a clarification question, format your question as: 
Question: """{{YOUR_QUESTION_HERE}}"""

To ask multiple questions in a single turn, you should list your questions as:
Question: """
1. {{First_Question}}
2. {{Second_Question}}
3. {{Third_Question}}
...
"""

If you are ready to write the final HTML code, format your code as
```
{{HTML_CSS_CODE}}
```
Remember to use "rick.jpg" as the placeholder for any images
\end{lstlisting}
\end{figure*}

\begin{figure*}
\begin{lstlisting}[caption={The user prompt used for HTML prototype generation augmented with question-answer pairs.}, label={lst:qa_augmented_prompt}, captionpos=b]
Here is a sketch design of a webpage drawn in the wireframing conventions. Also, here is a list of text blocks that I would like to include in the webpage:

{texts}

Could you write a HTML+CSS code of this webpage for me?

Here are some additional information for your reference:
{qa_pairs}

Please format your code as
```
{{HTML_CSS_CODE}}
```
Remember to use "rick.jpg" as the placeholder for any images
\end{lstlisting}
\end{figure*}

\begin{table*}[ht]
\small
\centering
\begin{tabular}{@{}ll|rrrrrr|rr@{}}
\toprule
\textbf{Metric} & \textbf{Model} & \multicolumn{6}{c|}{\textbf{Performance per Turn}} & \multicolumn{2}{c}{\textbf{Improv. per Turn}} \\
\cmidrule(lr){3-8} \cmidrule(lr){9-10}
 &  & \textbf{k=0} & \textbf{k=1} & \textbf{k=2} & \textbf{k=3} & \textbf{k=4} & \textbf{k=5} & \textbf{Avg}$\uparrow$ & \textbf{Std}$\downarrow$ \\
\midrule
Visual Similarity & \makecell{GPT-4O} & 82.29 & \cellcolor[rgb]{0.81, 0.93, 0.78} 84.28* & \cellcolor[rgb]{0.82, 0.93, 0.80} 85.24* & \cellcolor[rgb]{0.91, 0.96, 0.89} 85.67* & \cellcolor[rgb]{0.92, 0.97, 0.90} 86.10* & \cellcolor[rgb]{0.94, 0.98, 0.93} 86.29 & \textbf{0.80} & 4.75 \\
 & \makecell{Claude 3 Opus} & 81.75 & \cellcolor[rgb]{0.89, 0.96, 0.87} 82.93* & \cellcolor[rgb]{0.83, 0.93, 0.81} 83.87* & \cellcolor[rgb]{0.90, 0.96, 0.88} 84.37 & \cellcolor[rgb]{0.94, 0.98, 0.93} 84.60 & \cellcolor[rgb]{0.86, 0.95, 0.84} 85.18* & 0.63 & \textbf{2.59} \\
 & \makecell{Gemini 1.5 Pro} & 80.87 & \cellcolor[rgb]{0.79, 0.92, 0.77} 83.00 & \cellcolor[rgb]{0.95, 0.98, 0.94} 83.13* & 82.44 & \cellcolor[rgb]{0.92, 0.97, 0.90} 82.90 & \cellcolor[rgb]{0.88, 0.95, 0.86} 83.43* & 0.51 & 3.43 \\
\midrule
Block Match & \makecell{GPT-4O} & 84.74 & \cellcolor[rgb]{0.76, 0.90, 0.73} 87.21* & \cellcolor[rgb]{0.85, 0.94, 0.82} 88.07* & \cellcolor[rgb]{0.85, 0.94, 0.82} 88.77* & \cellcolor[rgb]{0.91, 0.97, 0.90} 89.24 & \cellcolor[rgb]{0.93, 0.97, 0.92} 89.47 & \textbf{0.95} & 9.77 \\
 & \makecell{Claude 3 Opus} & 80.83 & \cellcolor[rgb]{0.75, 0.90, 0.72} 83.33* & \cellcolor[rgb]{0.90, 0.96, 0.88} 83.89 & \cellcolor[rgb]{0.81, 0.93, 0.78} 84.74 & \cellcolor[rgb]{0.94, 0.98, 0.92} 85.03 & \cellcolor[rgb]{0.83, 0.93, 0.81} 85.74 & 0.86 & \textbf{6.10} \\
 & \makecell{Gemini 1.5 Pro} & 77.01 & \cellcolor[rgb]{0.57, 0.82, 0.56} 80.82* & 79.70 & 78.32 & \cellcolor[rgb]{0.45, 0.77, 0.46} 80.71 & \cellcolor[rgb]{0.86, 0.95, 0.84} 81.29 & 0.86 & 11.01 \\
\midrule
Text & \makecell{GPT-4O} & 97.25 & \cellcolor[rgb]{0.95, 0.98, 0.94} 97.57* & \cellcolor[rgb]{0.94, 0.98, 0.93} 97.80 & \cellcolor[rgb]{0.96, 0.99, 0.96} 97.84 & \cellcolor[rgb]{0.94, 0.98, 0.93} 98.05 & 97.90 & 0.13 & 4.15 \\
 & \makecell{Claude 3 Opus} & 96.85 & \cellcolor[rgb]{0.94, 0.98, 0.93} 97.25 & 97.18 & \cellcolor[rgb]{0.92, 0.97, 0.91} 97.50 & 97.44 & 97.30 & 0.05 & \textbf{1.87} \\
 & \makecell{Gemini 1.5 Pro} & 94.89 & \cellcolor[rgb]{0.70, 0.88, 0.67} 97.78* & \cellcolor[rgb]{0.96, 0.99, 0.95} 97.84 & 97.80 & 97.62 & \cellcolor[rgb]{0.96, 0.98, 0.95} 97.69 & \textbf{0.56} & 4.33 \\
\midrule
Position & \makecell{GPT-4O} & 76.83 & \cellcolor[rgb]{0.81, 0.93, 0.78} 78.82* & \cellcolor[rgb]{0.82, 0.93, 0.80} 79.80* & \cellcolor[rgb]{0.95, 0.98, 0.94} 79.93 & \cellcolor[rgb]{0.90, 0.96, 0.88} 80.52* & \cellcolor[rgb]{0.91, 0.97, 0.90} 80.87 & 0.81 & 6.75 \\
 & \makecell{Claude 3 Opus} & 75.88 & \cellcolor[rgb]{0.92, 0.97, 0.90} 76.70 & \cellcolor[rgb]{0.48, 0.78, 0.48} 78.97 & \cellcolor[rgb]{0.96, 0.99, 0.96} 79.01 & \cellcolor[rgb]{0.91, 0.96, 0.89} 79.53 & \cellcolor[rgb]{0.45, 0.77, 0.46} 81.32 & \textbf{1.03} & 7.11 \\
 & \makecell{Gemini 1.5 Pro} & 75.16 & \cellcolor[rgb]{0.66, 0.86, 0.64} 78.35* & \cellcolor[rgb]{0.75, 0.90, 0.72} 79.66 & 78.69 & \cellcolor[rgb]{0.96, 0.98, 0.95} 78.80 & \cellcolor[rgb]{0.82, 0.93, 0.80} 79.53 & 0.88 & \textbf{5.60} \\
\midrule
Color & \makecell{GPT-4O} & 65.46 & \cellcolor[rgb]{0.45, 0.77, 0.46} 70.00* & \cellcolor[rgb]{0.45, 0.77, 0.46} 72.36* & \cellcolor[rgb]{0.76, 0.90, 0.73} 73.41* & \cellcolor[rgb]{0.89, 0.96, 0.87} 74.03* & \cellcolor[rgb]{0.90, 0.96, 0.88} 74.48 & \textbf{1.80} & 10.86 \\
 & \makecell{Claude 3 Opus} & 68.66 & \cellcolor[rgb]{0.91, 0.97, 0.89} 69.61 & \cellcolor[rgb]{0.59, 0.83, 0.57} 71.51 & \cellcolor[rgb]{0.86, 0.95, 0.84} 72.16 & 71.87 & \cellcolor[rgb]{0.90, 0.96, 0.88} 72.31 & 0.66 & 5.97 \\
 & \makecell{Gemini 1.5 Pro} & 70.61 & 70.61 & \cellcolor[rgb]{0.92, 0.97, 0.90} 71.03 & 70.13 & 69.87 & \cellcolor[rgb]{0.77, 0.91, 0.74} 70.80 & 0.04 & \textbf{5.38} \\
\midrule
CLIP & \makecell{GPT-4O} & 87.16* & \cellcolor[rgb]{0.93, 0.97, 0.91} 87.80 & \cellcolor[rgb]{0.93, 0.97, 0.91} 88.15 & \cellcolor[rgb]{0.93, 0.97, 0.92} 88.41 & \cellcolor[rgb]{0.94, 0.98, 0.92} 88.68 & \cellcolor[rgb]{0.96, 0.99, 0.95} 88.73 & 0.31 & 4.47 \\
 & \makecell{Claude 3 Opus} & 86.52 & \cellcolor[rgb]{0.89, 0.96, 0.87} 87.78* & \cellcolor[rgb]{0.97, 0.99, 0.96} 87.80 & \cellcolor[rgb]{0.86, 0.94, 0.83} 88.46 & \cellcolor[rgb]{0.88, 0.95, 0.86} 89.14 & \cellcolor[rgb]{0.95, 0.98, 0.94} 89.24 & \textbf{0.52} & 3.51 \\
 & \makecell{Gemini 1.5 Pro} & 86.70 & \cellcolor[rgb]{0.92, 0.97, 0.91} 87.42 & \cellcolor[rgb]{0.97, 0.99, 0.96} 87.44 & 87.27 & \cellcolor[rgb]{0.95, 0.98, 0.93} 87.47 & \cellcolor[rgb]{0.91, 0.97, 0.90} 87.83 & 0.23 & \textbf{2.70} \\
\midrule
\midrule
Layout Similarity & \makecell{GPT-4O} & 20.38 & \cellcolor[rgb]{0.96, 0.98, 0.95} 20.56 & \cellcolor[rgb]{0.97, 0.99, 0.96} 20.57 & \cellcolor[rgb]{0.96, 0.99, 0.96} 20.61 & \cellcolor[rgb]{0.96, 0.99, 0.95} 20.67 & \cellcolor[rgb]{0.88, 0.95, 0.85} 21.21* & 0.17 & 4.89 \\
 & \makecell{Claude 3 Opus} & 17.11 & 16.97 & \cellcolor[rgb]{0.77, 0.91, 0.74} 18.20 & 18.20 & \cellcolor[rgb]{0.94, 0.98, 0.92} 18.47 & \cellcolor[rgb]{0.82, 0.93, 0.80} 19.20 & 0.39 & 5.40 \\
 & \makecell{Gemini 1.5 Pro} & 18.72 & \cellcolor[rgb]{0.93, 0.97, 0.91} 19.38 & \cellcolor[rgb]{0.84, 0.94, 0.82} 20.26 & \cellcolor[rgb]{0.94, 0.98, 0.93} 20.45 & \cellcolor[rgb]{0.90, 0.96, 0.89} 21.00 & \cellcolor[rgb]{0.90, 0.96, 0.88} 21.43 & \textbf{0.54} & \textbf{4.86} \\
\midrule
Text IoU & \makecell{GPT-4O} & 21.64 & \cellcolor[rgb]{0.95, 0.98, 0.94} 21.91 & 21.75 & 21.75 & \cellcolor[rgb]{0.94, 0.98, 0.93} 21.97 & \cellcolor[rgb]{0.89, 0.96, 0.87} 22.44* & 0.16 & \textbf{5.20} \\
 & \makecell{Claude 3 Opus} & 18.09 & \cellcolor[rgb]{0.92, 0.97, 0.90} 18.95 & \cellcolor[rgb]{0.81, 0.93, 0.78} 19.99 & 19.15 & \cellcolor[rgb]{0.75, 0.90, 0.72} 20.46 & \cellcolor[rgb]{0.65, 0.86, 0.62} 21.76* & \textbf{0.70} & 5.24 \\
 & \makecell{Gemini 1.5 Pro} & 19.46 & \cellcolor[rgb]{0.90, 0.96, 0.89} 20.49 & \cellcolor[rgb]{0.90, 0.96, 0.89} 21.04 & \cellcolor[rgb]{0.91, 0.97, 0.90} 21.42 & \cellcolor[rgb]{0.96, 0.98, 0.95} 21.53 & 21.44 & 0.40 & 5.31 \\
\midrule
Image IoU & \makecell{GPT-4O} & 13.61 & 13.26 & \cellcolor[rgb]{0.93, 0.97, 0.92} 13.56 & 13.26 & 13.17 & \cellcolor[rgb]{0.91, 0.96, 0.89} 13.56* & -0.01 & 6.40 \\
 & \makecell{Claude 3 Opus} & 8.32 & 6.62 & \cellcolor[rgb]{0.94, 0.98, 0.93} 6.84 & \cellcolor[rgb]{0.94, 0.98, 0.92} 7.08 & 5.79 & 5.76 & -0.51 & \textbf{5.93} \\
 & \makecell{Gemini 1.5 Pro} & 11.79 & \cellcolor[rgb]{0.96, 0.98, 0.95} 11.95 & \cellcolor[rgb]{0.89, 0.96, 0.87} 12.57 & 11.61 & \cellcolor[rgb]{0.71, 0.89, 0.69} 13.08* & \cellcolor[rgb]{0.86, 0.95, 0.84} 13.66 & \textbf{0.37} & 6.59 \\
\midrule
Other IoU & \makecell{GPT-4O} & 3.64 & \cellcolor[rgb]{0.96, 0.98, 0.95} 3.87 & \cellcolor[rgb]{0.95, 0.98, 0.94} 4.04 & \cellcolor[rgb]{0.96, 0.99, 0.95} 4.09 & 3.97 & \cellcolor[rgb]{0.94, 0.98, 0.93} 4.16 & 0.10 & \textbf{5.57} \\
 & \makecell{Claude 3 Opus} & 2.97 & \cellcolor[rgb]{0.89, 0.96, 0.87} 4.17 & \cellcolor[rgb]{0.96, 0.98, 0.95} 4.29 & \cellcolor[rgb]{0.45, 0.77, 0.46} 6.23* & 5.77 & 5.71 & 0.51 & 8.67 \\
 & \makecell{Gemini 1.5 Pro} & 0.96 & \cellcolor[rgb]{0.84, 0.94, 0.82} 2.63* & \cellcolor[rgb]{0.89, 0.96, 0.87} 3.24 & 2.40 & \cellcolor[rgb]{0.68, 0.87, 0.66} 3.99* & \cellcolor[rgb]{0.65, 0.86, 0.62} 5.29* & \textbf{0.86} & 5.80 \\
\bottomrule
\end{tabular}
\caption{Models’ performance per turn on the user feedback benchmark, where the simulated user compares the model’s implementation at each turn against a reference implementation and provides feedback instructions. Color intensity indicates the average improvement over the previous turn, and * indicates statistical significance (p<0.05).}
\label{tab:feedback}
\end{table*}

\begin{table*}[ht]
\small
\centering
\begin{tabular}{@{}ll|rrrrrr|rr@{}}
\toprule
\textbf{Metric} & \textbf{Model} & \multicolumn{6}{c|}{\textbf{Performance per Turn}} & \multicolumn{2}{c}{\textbf{Improv. per Turn}} \\
\cmidrule(lr){3-8} \cmidrule(lr){9-10}
 &  & \textbf{k=0} & \textbf{k=1} & \textbf{k=2} & \textbf{k=3} & \textbf{k=4} & \textbf{k=5} & \textbf{Avg}$\uparrow$ & \textbf{Std}$\downarrow$ \\
\midrule
Visual Similarity & \makecell{GPT-4O} & 81.03 & \cellcolor[rgb]{0.91, 0.96, 0.89} 81.53 & \cellcolor[rgb]{0.86, 0.95, 0.84} 82.82* & 82.30 & \cellcolor[rgb]{0.94, 0.98, 0.92} 82.55 & 82.36 & \textbf{0.27} & 3.31 \\
 & \makecell{Claude 3 Opus} & 81.29 & \cellcolor[rgb]{0.92, 0.97, 0.91} 81.69 & \cellcolor[rgb]{0.97, 0.99, 0.96} 81.75 & 81.64 & \cellcolor[rgb]{0.84, 0.94, 0.81} 82.50 & 82.50 & 0.24 & 3.63 \\
 & \makecell{Gemini 1.5 Pro} & 80.79 & \cellcolor[rgb]{0.93, 0.97, 0.92} 81.11 & 80.94 & \cellcolor[rgb]{0.93, 0.97, 0.91} 81.48 & 81.48 & 81.48 & 0.14 & \textbf{2.93} \\
\midrule
Block Match & \makecell{GPT-4O} & 75.10 & \cellcolor[rgb]{0.74, 0.90, 0.71} 76.44 & \cellcolor[rgb]{0.56, 0.82, 0.55} 79.76* & 77.19 & \cellcolor[rgb]{0.78, 0.91, 0.75} 78.34 & \cellcolor[rgb]{0.93, 0.97, 0.92} 78.48 & \textbf{0.68} & 9.37 \\
 & \makecell{Claude 3 Opus} & 79.36 & \cellcolor[rgb]{0.46, 0.77, 0.47} 81.67 & 80.42 & 79.88 & \cellcolor[rgb]{0.45, 0.77, 0.46} 82.16 & 80.74 & 0.28 & 14.26 \\
 & \makecell{Gemini 1.5 Pro} & 76.02 & \cellcolor[rgb]{0.97, 0.99, 0.96} 76.03 & 73.36 & \cellcolor[rgb]{0.45, 0.77, 0.46} 77.14* & 77.14 & 77.14 & 0.22 & \textbf{9.21} \\
\midrule
Text & \makecell{GPT-4O} & 96.74 & \cellcolor[rgb]{0.93, 0.97, 0.91} 97.09 & \cellcolor[rgb]{0.93, 0.97, 0.92} 97.61 & 96.88 & \cellcolor[rgb]{0.94, 0.98, 0.93} 97.12 & \cellcolor[rgb]{0.94, 0.98, 0.93} 97.21 & 0.09 & \textbf{2.86} \\
 & \makecell{Claude 3 Opus} & 97.27 & 96.94 & \cellcolor[rgb]{0.96, 0.99, 0.95} 97.05 & \cellcolor[rgb]{0.96, 0.99, 0.95} 97.14 & 97.06 & \cellcolor[rgb]{0.64, 0.86, 0.62} 97.81* & \textbf{0.11} & 2.93 \\
 & \makecell{Gemini 1.5 Pro} & 96.19 & 95.81 & \cellcolor[rgb]{0.92, 0.97, 0.91} 96.43 & 95.91 & 95.91 & 95.91 & -0.06 & 3.16 \\
\midrule
Position & \makecell{GPT-4O} & 76.37 & \cellcolor[rgb]{0.83, 0.94, 0.81} 77.28 & \cellcolor[rgb]{0.69, 0.87, 0.66} 79.85* & 79.13 & \cellcolor[rgb]{0.93, 0.97, 0.91} 79.46 & 78.36 & \textbf{0.40} & 8.78 \\
 & \makecell{Claude 3 Opus} & 75.62 & 75.39 & \cellcolor[rgb]{0.93, 0.97, 0.91} 75.98 & 75.37 & \cellcolor[rgb]{0.91, 0.97, 0.90} 75.83 & \cellcolor[rgb]{0.45, 0.77, 0.46} 76.86 & 0.25 & 8.92 \\
 & \makecell{Gemini 1.5 Pro} & 74.44 & \cellcolor[rgb]{0.93, 0.97, 0.91} 74.77 & \cellcolor[rgb]{0.81, 0.93, 0.79} 76.47 & 74.91 & 74.91 & 74.91 & 0.09 & \textbf{7.10} \\
\midrule
Color & \makecell{GPT-4O} & 69.94 & \cellcolor[rgb]{0.94, 0.98, 0.93} 70.16 & 69.60 & \cellcolor[rgb]{0.82, 0.93, 0.80} 71.16 & 70.47 & \cellcolor[rgb]{0.95, 0.98, 0.94} 70.53 & 0.12 & 8.25 \\
 & \makecell{Claude 3 Opus} & 68.51 & \cellcolor[rgb]{0.94, 0.98, 0.93} 68.74 & \cellcolor[rgb]{0.96, 0.99, 0.96} 68.83 & \cellcolor[rgb]{0.89, 0.96, 0.87} 69.87 & \cellcolor[rgb]{0.54, 0.81, 0.53} 71.87 & 70.93 & \textbf{0.48} & 8.05 \\
 & \makecell{Gemini 1.5 Pro} & 70.49 & \cellcolor[rgb]{0.70, 0.88, 0.67} 72.00 & \cellcolor[rgb]{0.96, 0.98, 0.95} 72.16 & \cellcolor[rgb]{0.93, 0.97, 0.91} 72.74 & 72.74 & 72.74 & 0.45 & \textbf{5.87} \\
\midrule
CLIP & \makecell{GPT-4O} & 86.99 & 86.68 & \cellcolor[rgb]{0.92, 0.97, 0.91} 87.30 & 87.17 & \cellcolor[rgb]{0.95, 0.98, 0.93} 87.36 & 87.21 & 0.04 & 3.35 \\
 & \makecell{Claude 3 Opus} & 85.68 & \cellcolor[rgb]{0.96, 0.99, 0.96} 85.72 & \cellcolor[rgb]{0.92, 0.97, 0.90} 86.46 & 85.94 & 85.57 & \cellcolor[rgb]{0.73, 0.89, 0.71} 86.17 & \textbf{0.10} & 3.90 \\
 & \makecell{Gemini 1.5 Pro} & 86.80 & \cellcolor[rgb]{0.96, 0.98, 0.95} 86.92 & 86.29 & \cellcolor[rgb]{0.94, 0.98, 0.93} 86.69 & 86.69 & 86.69 & -0.02 & \textbf{2.65} \\
\midrule
\midrule
Layout Similarity & \makecell{GPT-4O} & 21.33 & \cellcolor[rgb]{0.93, 0.97, 0.91} 21.68 & \cellcolor[rgb]{0.95, 0.98, 0.94} 21.93 & 21.08 & \cellcolor[rgb]{0.91, 0.96, 0.89} 21.58 & 20.93 & -0.08 & 7.85 \\
 & \makecell{Claude 3 Opus} & 16.46 & 15.33 & \cellcolor[rgb]{0.45, 0.77, 0.46} 19.23* & 18.08 & \cellcolor[rgb]{0.86, 0.94, 0.83} 18.86 & 15.33 & -0.23 & 11.57 \\
 & \makecell{Gemini 1.5 Pro} & 18.29 & \cellcolor[rgb]{0.83, 0.93, 0.80} 19.22* & 18.09 & \cellcolor[rgb]{0.93, 0.97, 0.92} 18.59 & 18.59 & 18.59 & \textbf{0.06} & \textbf{7.25} \\
\midrule
Text IoU & \makecell{GPT-4O} & 22.08 & 21.57 & \cellcolor[rgb]{0.90, 0.96, 0.88} 22.49 & 21.90 & \cellcolor[rgb]{0.90, 0.96, 0.88} 22.45 & 20.97 & -0.22 & 8.03 \\
 & \makecell{Claude 3 Opus} & 18.02 & 17.21 & \cellcolor[rgb]{0.75, 0.90, 0.72} 19.39 & 18.14 & \cellcolor[rgb]{0.86, 0.95, 0.84} 18.90 & 17.91 & -0.02 & 9.66 \\
 & \makecell{Gemini 1.5 Pro} & 19.26 & \cellcolor[rgb]{0.45, 0.77, 0.46} 21.59* & 20.64 & 20.49 & 20.49 & 20.49 & \textbf{0.25} & \textbf{7.10} \\
\midrule
Image IoU & \makecell{GPT-4O} & 13.23 & \cellcolor[rgb]{0.70, 0.88, 0.67} 14.74 & 14.25 & 12.82 & \cellcolor[rgb]{0.85, 0.94, 0.83} 13.61 & 12.88 & -0.07 & 9.93 \\
 & \makecell{Claude 3 Opus} & 9.11 & 9.06 & \cellcolor[rgb]{0.79, 0.92, 0.76} 10.92 & \cellcolor[rgb]{0.87, 0.95, 0.85} 12.10 & 10.22 & 6.32 & -0.56 & 13.75 \\
 & \makecell{Gemini 1.5 Pro} & 11.19 & 10.20 & 9.56 & \cellcolor[rgb]{0.78, 0.91, 0.75} 11.46 & 11.46 & 11.46 & \textbf{0.05} & \textbf{8.60} \\
\midrule
Other IoU & \makecell{GPT-4O} & 2.75 & \cellcolor[rgb]{0.93, 0.97, 0.91} 3.10 & 2.87 & \cellcolor[rgb]{0.77, 0.91, 0.74} 4.83 & 1.67 & \cellcolor[rgb]{0.94, 0.98, 0.93} 1.76 & -0.20 & 9.32 \\
 & \makecell{Claude 3 Opus} & 2.22 & 1.78 & \cellcolor[rgb]{0.91, 0.97, 0.89} 2.60 & 2.48 & \cellcolor[rgb]{0.85, 0.94, 0.82} 3.31 & 2.91 & \textbf{0.14} & 7.58 \\
 & \makecell{Gemini 1.5 Pro} & 0.74 & \cellcolor[rgb]{0.91, 0.96, 0.89} 1.25 & \cellcolor[rgb]{0.95, 0.98, 0.94} 1.50 & 0.67 & 0.67 & 0.67 & -0.01 & \textbf{3.81} \\
\bottomrule
\end{tabular}
\caption{Models' performance per turn on the question asking benchmark, where the model proactively asks a question about the sketch at each turn, and generate the HTML code based on the answer from the simulated user. Color intensity indicates the average amount of improvement over the previous turn, and * indicates statistical significance (p<0.05).}
\label{tab:qa}
\end{table*}

\begin{table*}[ht]
\centering
\begin{tabular}{@{}l l | c c@{}}
\toprule
\textbf{Task} & \textbf{Models} & \textbf{Avg $\Delta$ Visual Score $\downarrow$} & \textbf{Avg $\Delta$ Layout Score $\downarrow$} \\ 
\midrule
\multirow{10}{*}{Feedback Following}
    & GPT-4O & 0.04227 & 0.08912 \\
    & GPT-4O Mini & \textbf{0.03794} & \textbf{0.05563} \\
    & Claude 3.5 Sonnet & 0.09638 & 0.08418 \\
    & Claude 3 Opus & 0.05141 & 0.07544 \\
    & Claude 3 Sonnet & 0.05678 & 0.13398 \\
    & Claude 3 Haiku & 0.09947 & 0.08634 \\
    & Gemini 1.5 Pro & 0.04464 & 0.14815 \\
    & Gemini 1.5 Flash & 0.28172 & 0.12727 \\
    & InternVL2-8b & 0.28646 & 0.10428 \\
    & Llava-1.6-8b & 0.56875 & 0.11680 \\
\midrule
\multirow{10}{*}{Question Asking}
    & GPT-4O & 0.04470 & 0.10609 \\
    & GPT-4O Mini & 0.05444 & 0.11776 \\
    & Claude 3.5 Sonnet & 0.04480 & 0.11098 \\
    & Claude 3 Opus & \textbf{0.04143} & 0.11039 \\
    & Claude 3 Sonnet & 0.06045 & 0.10629 \\
    & Claude 3 Haiku & 0.06298 & 0.13278 \\
    & Gemini 1.5 Pro & 0.04420 & \textbf{0.09528} \\
    & Gemini 1.5 Flash & 0.13824 & 0.11112 \\
    & InternVL2-8b* & - & - \\
    & Llava-1.6-8b* & - & - \\
\midrule
\multirow{10}{*}{Direct Generation}
    & GPT-4O & 0.11307 & - \\
    & GPT-4O Mini & 0.07477 & - \\
    & Claude 3.5 Sonnet & 0.10815 & - \\
    & Claude 3 Opus & \textbf{0.06886} & - \\
    & Claude 3 Sonnet & 0.12071 & - \\
    & Claude 3 Haiku & 0.06121 & - \\
    & Gemini 1.5 Pro & 0.10561 & - \\
    & Gemini 1.5 Flash & 0.13244 & - \\
    & InternVL2-8b & 0.11371 & - \\
    & Llava-1.6-8b & 0.09335 & - \\
\bottomrule
\end{tabular}
\caption{Mean delta performance scores across the three evaluation tasks. *Both InternVL2-8b and Llava-1.6-8b failed to generate valid HTML outputs in the multi-turn question-asking scenario, and thus they are not given the delta scores.}
\label{tab:sketch_variation}
\end{table*}

\section{Additional Experiment Results}
\label{appendix:additional_experiment_results}

\paragraph{Direct Evaluation}
With the additional context from text-augmented prompting, it is frequently observed that the two open-source models would either repeat parts of the generated code until max tokens exceeded, or output the <EOS> token prematurely before the HTML code is complete.

One may notice that the Text IoU scores are generally higher than Image IoU scores, which are higher than the IoUs of other tertiary components. This is because the IoU scores of a certain type of visual component is usually correlated to the corresponding area that type of components span across the webpage. Since text blocks take the largest areas in many webpages collected in the Sketch2Code dataset, models generally get the best scores in Text IoU. Contrarily, tertiary items such as buttons, search bars, and navigation menus only take a small area on the UI, it is especially challenging to achieve a high score on Other IoU.

\paragraph{Multi-turn Evaluation} We benchmarked the ten models on both multi-turn evaluation tasks with a maximum of five rounds of user interactions for each sketch. The two open-source models fail to operate under multi-turn interaction. Neither model could reliably generate correctly formatted questions, and the additional information provided by the simulated user often cause the model to degenerate. We present two failure case examples of open-source models in Appendix~\ref{appendix:qualitative_analysis}.

\begin{figure*}[ht]
\centering
\includegraphics[width=\textwidth]{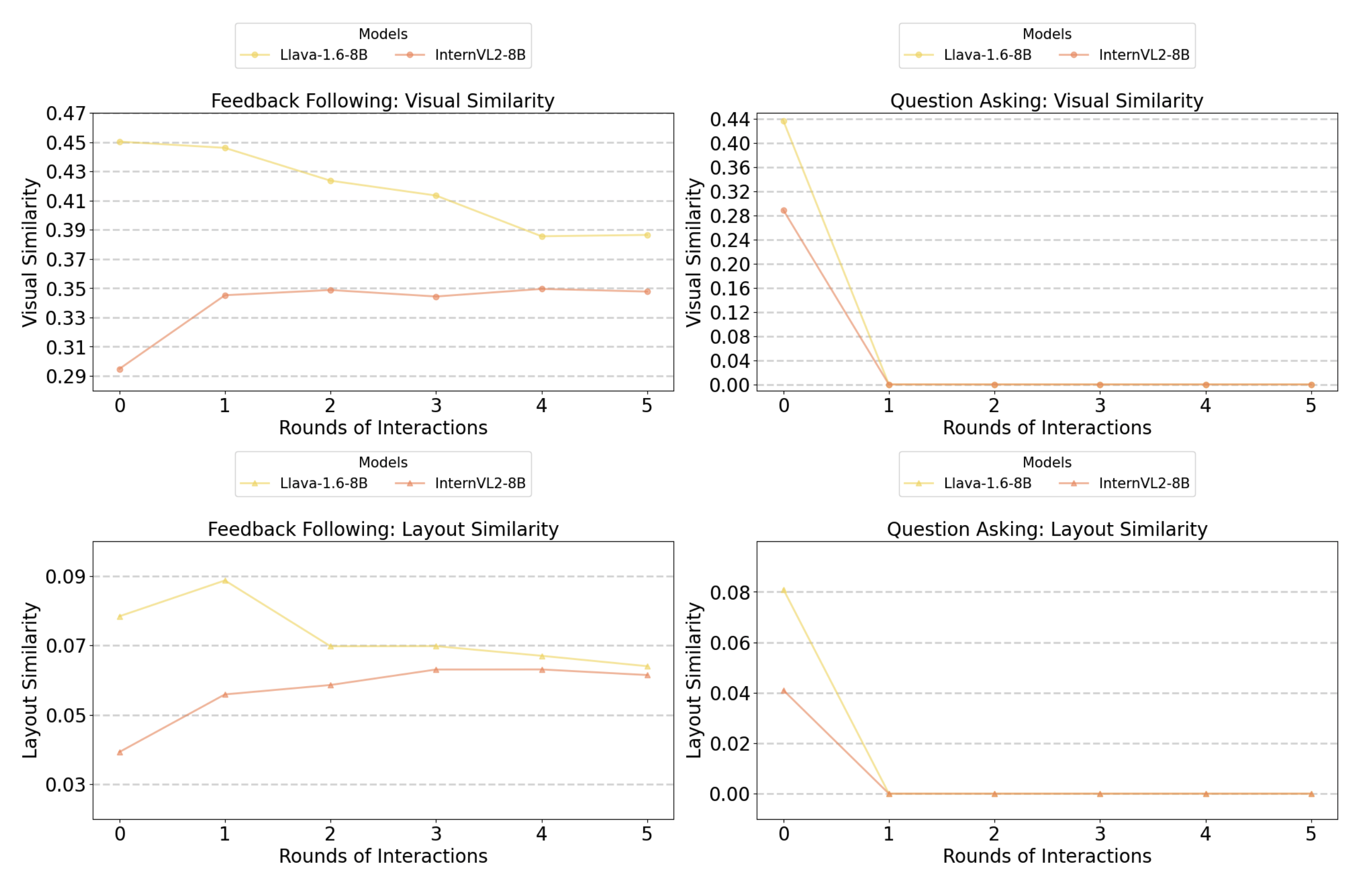}
\caption{The performances of open-source models on the feedback following benchmark (left) and the question asking benchmark (right).}
%\vspace{-1.5em}
\label{fig:performance_opensource}
\end{figure*}

\begin{figure*}[ht]
\centering
\includegraphics[width=\textwidth]{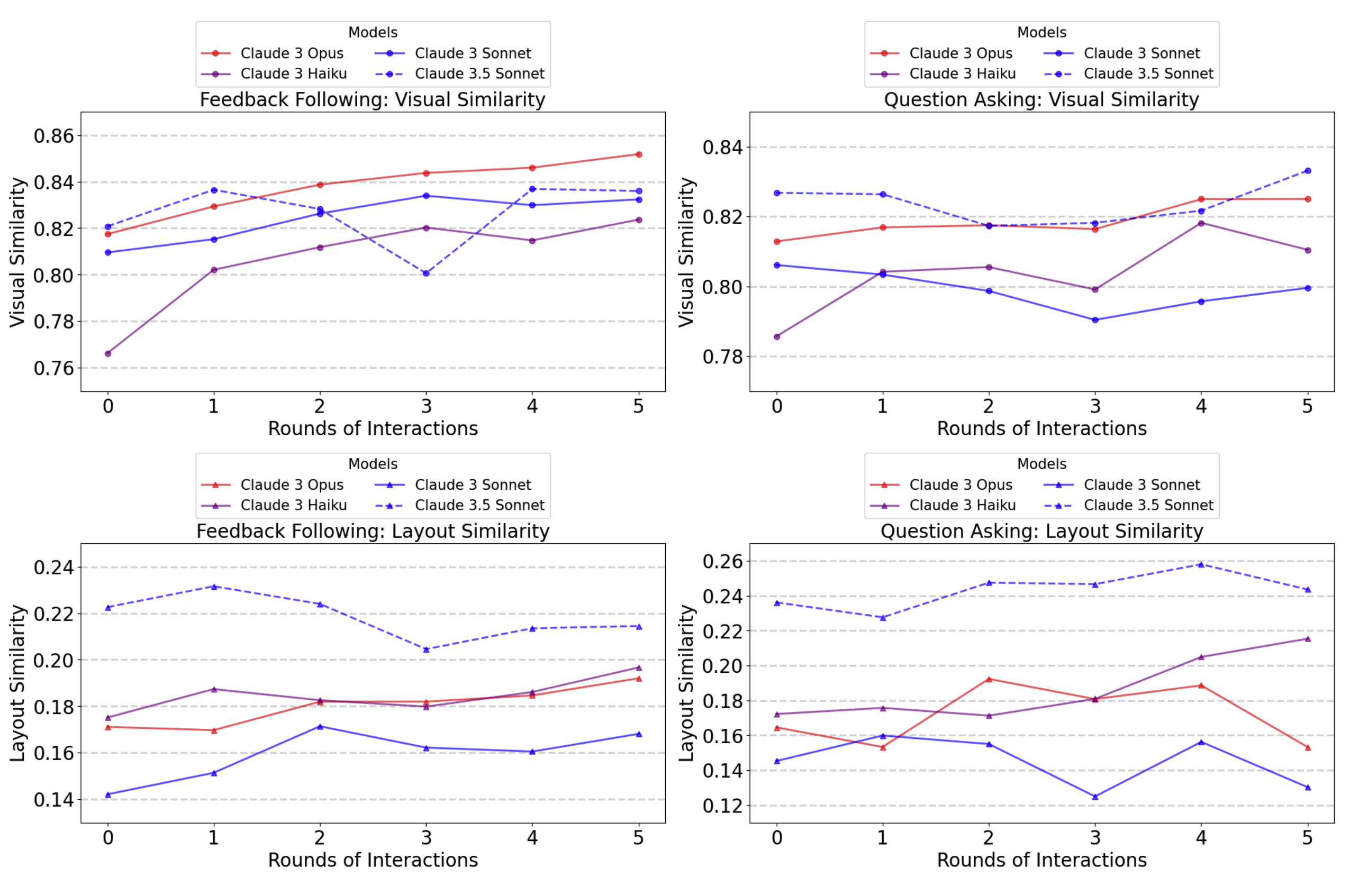}
\caption{The performances of the Claude 3 model family on the feedback following benchmark (left) and the question asking benchmark (right).}
%\vspace{-1.5em}
\label{fig:performance_claude}
\end{figure*}

Tables \ref{tab:feedback} and \ref{tab:qa} present detailed scoring breakdowns of GPT-4o, Claude 3 Opus, and Gemini 1.5 Pro on the two multi-turn benchmarks. We found that Gemini 1.5 Pro was rather reluctant to ask questions, often prematurely stopping before reaching the maximum five-question limit. Figure \ref{fig:performance_opensource} shows the performance of the two open-source models (Llava-1.6-8b and InternVL2-8b) on the multi-turn evaluation tasks. Figure~\ref{fig:performance_claude} shows the performances of the four models in the Claude 3 model family.

Table~\ref{tab:sketch_variation} shows the sensitivity of each model w.r.t. different variations of sketches. To examine the sensitivity to sketch variation, we calculated the average fluctuations a model has on different sketches drawn for the same webpages. The fluctuation of a performance metric is calculated as the max score minus the min score of outputs generated from all different sketch variations of the same webpage. For direct generation (with direct prompting), we computed the average fluctuation of layout scores per model. For the multi-turn interactive tasks, we calculated the fluctuation of both layout \& layout scores in the final turn of each generation.

\begin{figure*}
\begin{lstlisting}[caption={Instructions given to Prolific participants for pairwise layout similarity comparison}, label={lst:pairwise_sim_instructions}, captionpos=b]
Task Overview
In this survey, you will be given a reference webpage's screenshot, as well as two
candidate webpages (Example 1 and Example 2) that try to replicate the layout reference
webpage. Your task is to judge which of the two candidates has closer layout to the reference. Each (Reference, Example 1, Example 2) is presented in a row, where the original
boundary of screenshot is marked by black.

Note: All images in the original webpages are replaced by blue rectangles as placeholders. Sometimes, the candidate webpages will use "lorem ipsum..." as placeholder texts. Please disregard these placeholders and treat them as normal texts instead.

Comparison Guide
In this survey, you should base your comparisons solely on the layout similarities between the example webpages and the reference webpage.

You should pay close attention to:
Visual component matching: a good candidate example should contain all visual components that are present in the reference image and have no extra components. You should examine if the visual components in the candidate example match with the ones in the reference image. Visual components include text blocks, images, menus/navigation bars, tables, form inputs, etc.
Overall layout arrangement: a good candidate example should have its visual components arranged similarly as the reference image. You should examine how much the layout arrangement in each candidate overlap with the reference layout.
Size and position of each visual component: you should pay attention to the relative size and position of each visual component w.r.t. the overall width and height of the webpage.
You should ignore/not pay attention to:
Detailed text content: since the survey focuses solely on layout similarities, the exact details of textual content does not matter and should be factored out from your judgement. You should never rate a candidate example down if it contains placeholder texts, as long as text blocks are placed similarly as the reference image.
Color and Styles: background color, the color of each section/text block, the font/size/color of the text, or any other color \& styling decisions should never impact your final choice. Your decisions should be based on solely the layout and placement of each component.
\end{lstlisting}
\end{figure*}

\begin{figure*}
\begin{lstlisting}[caption={Instructions given to Prolific participants for simulated question answering annotation}, label={lst:sim_qa_instructions}, captionpos=b]
**Task Overview**

In this survey, you will be given pairs of webpage sketches and actual webpage screenshots. You will also be given a single turn of AI-generated conversation regarding the sketch. The conversation will include a question about the sketch generated by a simulated AI agent, along with an answer to that question generated by another simulated agent. You will then be asked to rate the quality of the question and answer.

To evaluate the quality of the generated question, you will be asked to rate its
1. Specificity: whether the question is relevant to specific parts or components of the sketch design.
2. Effectiveness: whether the question is effective at helping the agent better understand the intended layout and design of the webpage.
To evaluate the quality of the provided answer, you should focus on rating
1. Relevance: whether the provided answer directly responds to the given question.
2. Accuracy: whether the provided answer is accurate and faithful to the visual appearance of the reference webpage screenshot.
\end{lstlisting}
\end{figure*}

\begin{figure*}
\begin{lstlisting}[caption={Instructions given to Prolific participants for simulated user feedback annotation}, label={lst:sim_feedback_instructions}, captionpos=b]
**Task Overview**

In this survey, you will be asked to evaluate the quality of AI-simulated feedback for webpage implementation. In each question, you will be given a screenshot of the current implementation of a webpage and a screenshot of the intended (reference) implementation. You will then be provided with simulated feedback & instructions to help improve the current implementation. Your job is to evaluate the provided feedback & instructions via the following criteria:
1. Readability: the provided feedback & instructions should be clear and easy to follow.
2. Effectiveness: the provided feedback should effectively points out the visual difference between the current and reference implementations
\end{lstlisting}
\end{figure*}

\section{Additional Details on Human Annotation}
\label{appendix:human_annotation}
We recruit Prolific crowdworkers with an hourly rate of \$16. Participant are filtered based on their fluency in English, past Prolific participation (completed 2500+ surveys), and acceptance rate (98\%+). We recruited three qualified participants for each study and took the majority vote of the remaining three participants as the final label. Participants are asked for consent to share the annotation data and are given the choice to opt-out from the study. Identifying code leakage was a more challenging task that requires familiarity with coding and front-end engineering. In that regard, the first two authors conducted annotations on code leakage in 60 data samples.

For pairwise layout similarity annotation, participants are also given "Tie" as an option to handle the cases where making a comparison is difficult. However, these "Tie" cases are ignored during agreement calculation as it is hard to define a boundary for "Tie" using the automated metrics. Listing~\ref{lst:pairwise_sim_instructions} contains the full instructions given to participants for layout similarity annotation.

When annotating the simulated QA conversations, participants are presented with a sketch
design and a reference webpage, followed by a single round of simulated QA. The participants are then asked to evaluate the quality of the simulated user's answer based on the following two metrics: 1. Does the answer provided by the simulated user directly respond to the given question? 2. Does the answer from the simulated user stay faithful to the visual appearance of the reference webpage? When annotating code leakages among simulated answers, the two authors each independently labeled 60 examples. An answer is considered to have code leakage as long as one of the two authors responds "yes" to this question. The Fleiss-Kappa inter-annotator agreement among the prolific annotators for this task is 0.28. Listing~\ref{lst:sim_qa_instructions} shows the exact instructions given to participants for the simulated QA annotation.

For the simulated user feedback, participants are presented with a current implementation from a model, a reference implementation, and the feedback generated by the simulated user. Then, they are asked to answer the following two questions: 1. Are the provided instructions readable and easy to follow? 2. Does the simulated feedback accurately point out the visual difference between the current implementation and the reference webpage? The Fleiss Kappa score among the annotators in this task is 0.57. Listing~\ref{lst:sim_feedback_instructions} shows the instructions used for simulated feedback annotations.

\section{User Study Details}
\label{appendix:user_study}

The participating UI/UX practitioners are filtered on their knowledge and past projects with UI/UX design, familarity with fundamental sketching and wireframing concepts, fluency in English, as well as user ratings and job completion rates in the Upwork community. We compensated the interviewees with \$40-60 per hour. The exact amount of compensation was negotiated with each participant. We asked the participants for consent to record the interviews and share user study data before officially starting the contracts.

Among the eight selected participants, seven of them are based in the United States. Five of them have worked fulltime in tech companies in addition to freelancing, and three of them have worked only as freelancers. All participants have had more than 3 years of industrial UI/UX design experience, and seven of them have more than 5 years of experience.

During the user study, we asked each participant to interact with a demo sketch2code agent (using GPT-4o as the backbone VLM) and then invited them to provide feedback on the experience. We present our main user study findings below:

\paragraph{O1: Low-fidelity sketches play a substantial role in modern UI/UX development.}
All eight experts disclosed that they engaged with low-fidelity sketches in their past projects and that sketching plays a non-negligible role in the UI/UX design loop. The participants claimed various use cases for sketches, including communicating initial ideas and requirements with clients, collaborating with other designers, and working with early-stage ideas before switching to high-fidelity designs. Some interviewees pointed out that \textbf{sketching is especially important in research and large-scale projects, where the end goal is not clear from the beginning.} In these cases, sketches provide a fast and easy way to explore and iterate on design choices with little to no cost.

\paragraph{O2: Sketch2code agents can significantly accelerate the development cycle and lower communication barriers.}
All eight experts unanimously agreed that a sketch2code agent would significantly benefit their work. A sketch2code agent can help users easily flesh out early-stage ideas and brings down the communication barrier between clients and designers. One participant mentioned that they used to first educate the clients on common wireframing conventions in order to discuss the design requirements seamlessly. With sketch2code, the designer (or the client) can first play around with the agent to try out their ideas, and then directly use the generated prototype to communicate with their work partners.

Two other participant acknowledged that the sketch2code agent were able to visualize their ideas within a couple of minutes, while it would take hours for the participant to manually implement them in Figma themselves. Furthermore, faster visualization leads to faster feedback loop, and thus enabling researchers \& designers to explore more (initial) designs with lower cost, which is of vital importance for large-scale projects. Finally, a UX designer pointed out that \textit{"since the agent can lift the height of UI designs, I can focus more on studying the user experience."}

\paragraph{O3: Proactively asking questions is more helpful than passively following instructions.}
Even though existing models are more suited for the feedback-following mode, seven out of eight participants showed \textbf{strong preference} towards the question-asking agent. The interviewees expressed that they felt the need to specify every design detail to an agent that passively follows user instructions, whereas a question-asking agent can take over most of the cognitive workload, and the user would only need to focus on the parts that are being asked. Five participants claimed that the agent can sometimes catch the details and ambiguities on the sketch that the user would have otherwise missed. Three of the experts pointed out that \textbf{the agent can proactively “guide” the user through certain design decisions \& choices via a series of targeted questions}, so the user do not have to figure out every single detail themselves.

\paragraph{O4: Specifying Visual Components with Natural Language Alone is Challenging}
Like most existing VLM applications, the sketch2code agent uses a natural language interface. However, the participants pointed out that it is difficult to select visual components and specify visual details with natural language alone. When communicating design ideas in real life, people can simply point to specific visual components by mouse or by finger, and the communication is never based on language alone. Six out of the eight interview participants wished to have faster, easier, and more reliable ways to control the generated prototype (such as selecting and editing certian visual components with simple mouse clicks). One of the participants suggested that styles and color schemes can sometimes be hard to specify in natural language too. It would be more helpful if the user can upload custom style or color samples (e.g., a "mood board") for the agent to follow.

\section{Evaluating the Different Types of Generated Questions and Feedback}
\label{appendix:question_classification}

\begin{figure*}[!ht]
\centering
\includegraphics[width=\textwidth]{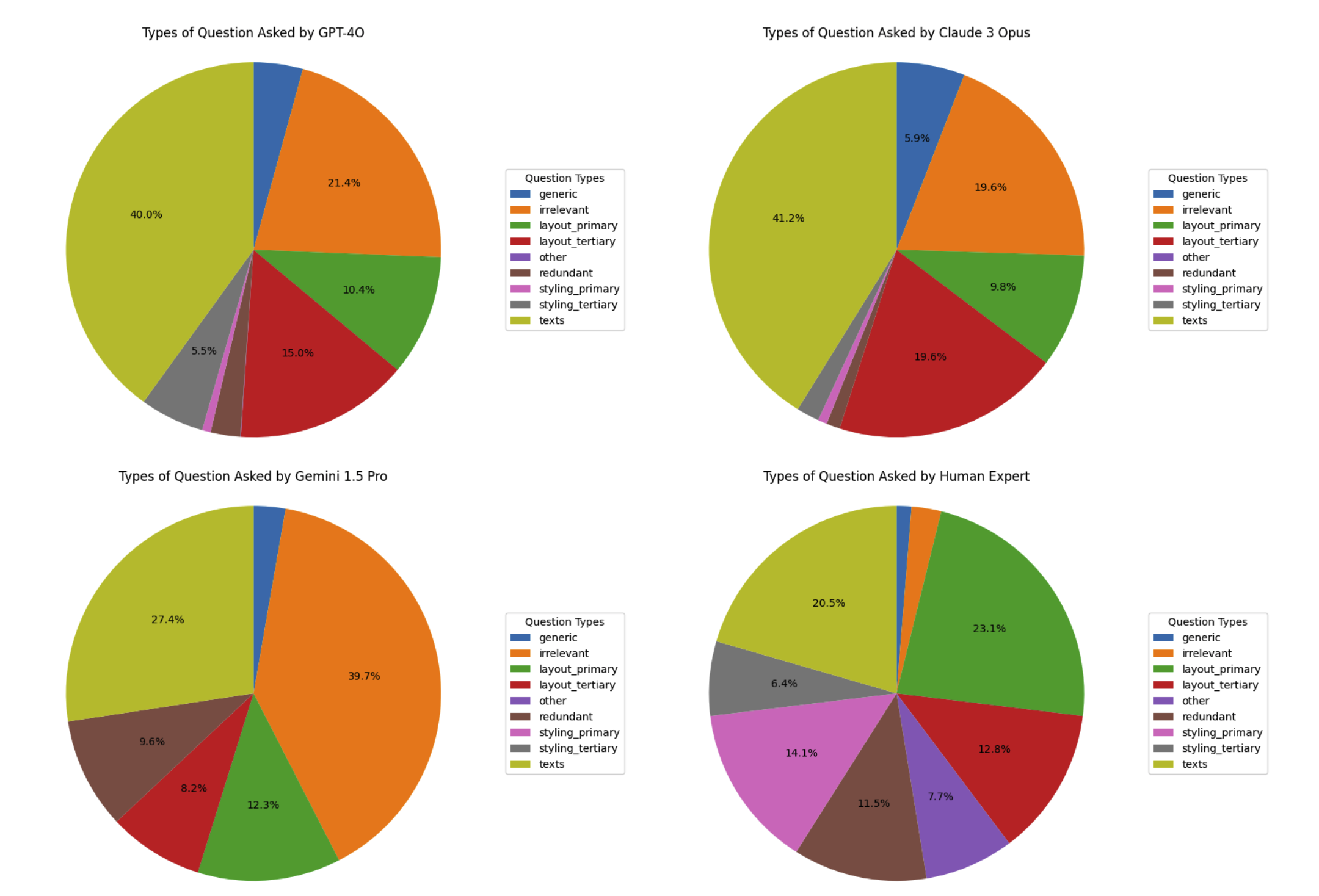}
\caption{Types of questions asked by different models and human experts: GPT-4o (top left), Claude 3 Opus (top right), Gemini 1.5 Pro (bottom left), Human Expert (bottom right). Zoom in for best view.}
%\vspace{-1.5em}
\label{fig:question_types}
\end{figure*}

\begin{figure*}[!ht]
\centering
\includegraphics[width=\textwidth]{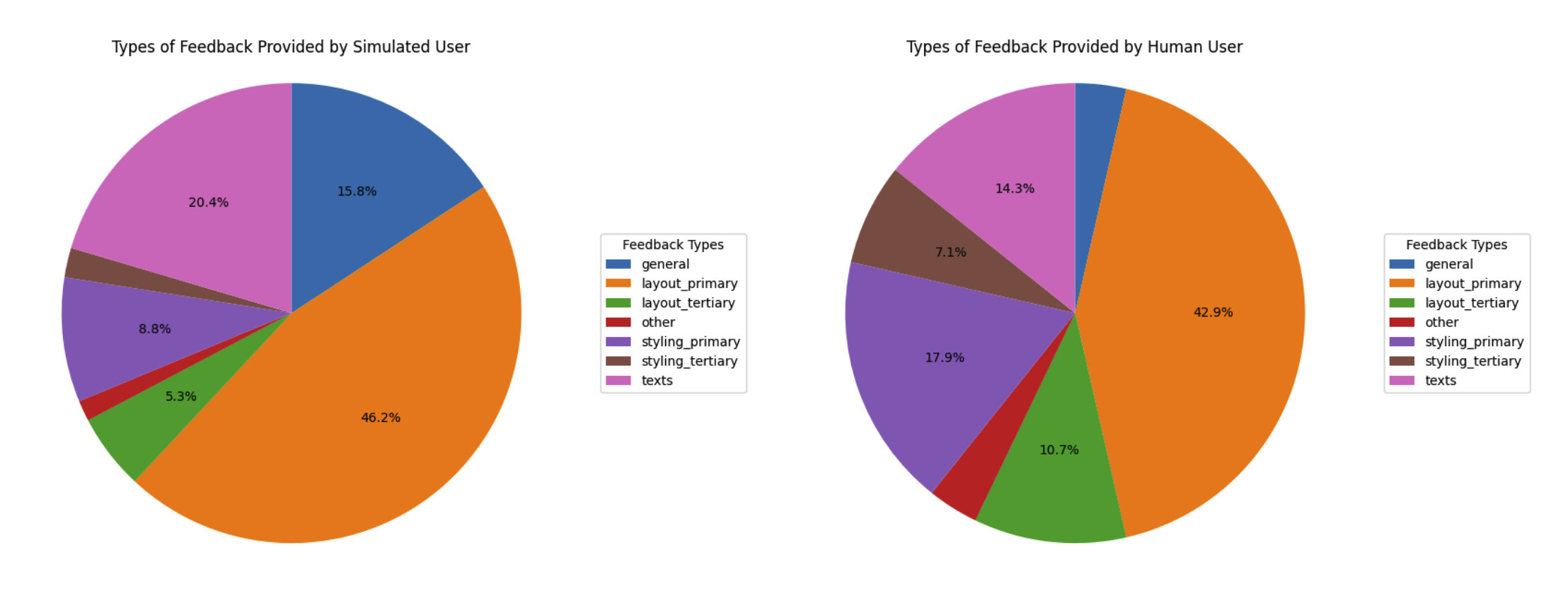}
\caption{Types of feedback provided by Simulated User (left) v.s. Human User (right). Zoom in for best view.}
% %\vspace{-1.5em}
\label{fig:feedback_types}
\end{figure*}

\begin{figure*}[ht]
\centering
\includegraphics[width=\textwidth]{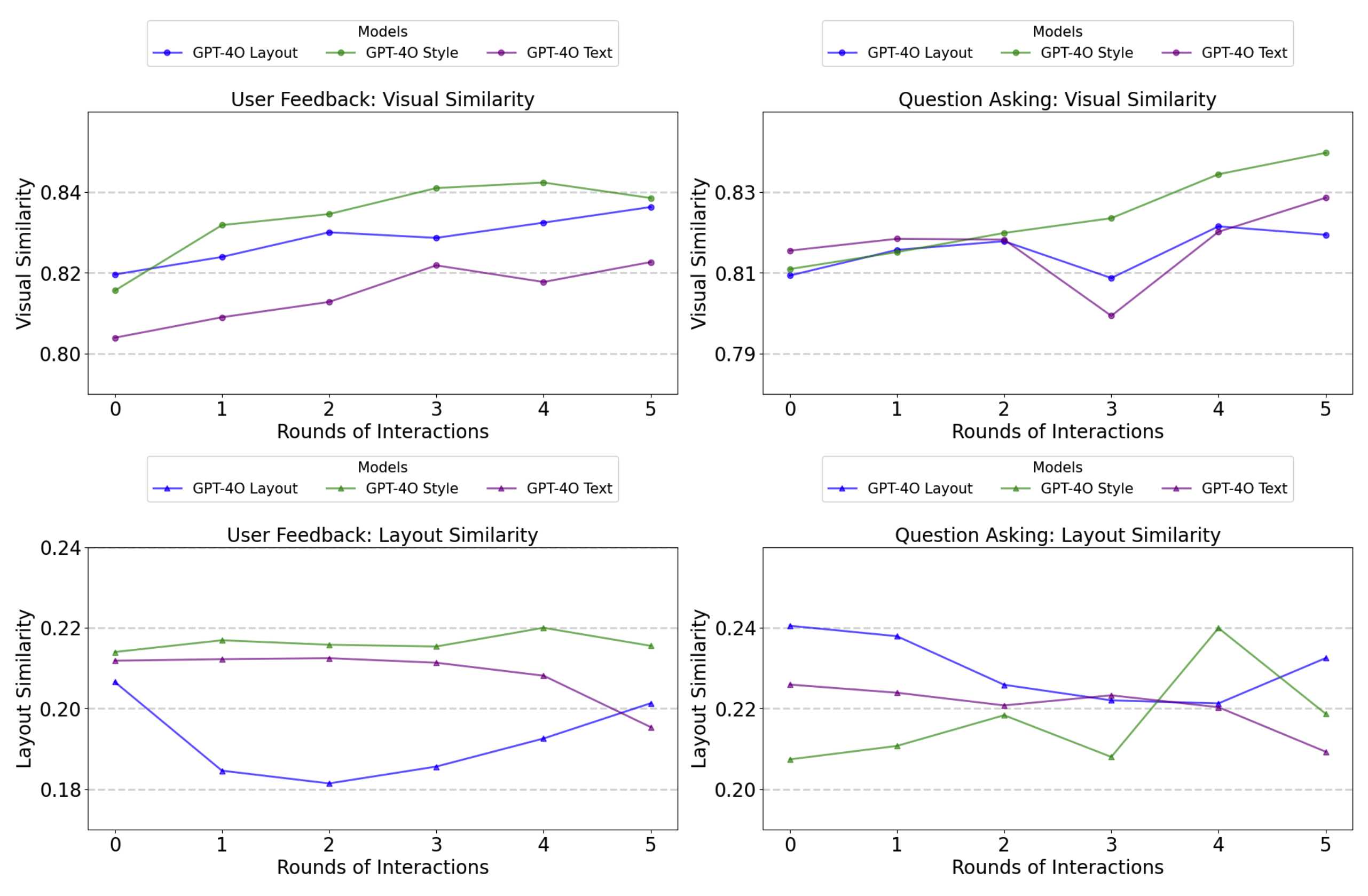}
\caption{Performance of GPT-4o on the question-asking benchmark when guided to prioritize questions about layout, styling, and text contents.}
\label{fig:guided_questions}
\end{figure*}

To evaluate the various questions asked and feedback given by sketch2code agents. We first performed HAC on the generated questions/feedback using cosine similarity of SBERT embeddings \citep{reimers2019sentencebert} with a similarity threshold of 0.6. For the ease of viewing, we then summarized the questions/feedback within each cluster through a GPT-4o summarizer. We manually looked through all clusters and extracted out the most common types of questions/feedback into a taxonomy. And finally, we classified each individual questions/feedback to one of the types in the taxonomy with GPT-4o. The full taxonomy for questions is available in Table \ref{tab:question_taxonomy}, and the taxonomy for feedback is available in Table \ref{tab:feedback_taxonomy}. Figures \ref{fig:question_types} and \ref{fig:feedback_types} shows the distribution of different types of questions and feedback, and Tables \ref{tab:question_improv_by_type} and \ref{tab:feedback_improv_by_type} shows the average improvement in visual and layout scores per model per question/feedback type. Figure \ref{fig:guided_questions} shows the performance of GPT-4o when guided to prioritize asking different types of questions.

\begin{table*}[!ht]
\centering
\begin{tabularx}{\textwidth}{@{}l|X@{}}
\toprule
\textbf{Question Type} & \textbf{Definition} \\ 
\midrule
Texts          & The question is asking about font size, font styles, and exact text content of textual components. \\
\midrule
Layout-Primary     & The question is asking about the overall webpage layout or the block size and positional placement of primary visual components. \\
\midrule
Layout-Tertiary        & The question is asking about the block size and positional placement of tertiary visual components. \\
\midrule
Styling-Primary     & The question is asking about the overall styling of the webpage or the color and styles of primary visual components \\
\midrule
Styling-Tertiary      & The question is asking about the color schemes, borders, and other stylistic characteristics of tertiary visual components. \\
\midrule
Generic & The question is too vague or generic. \\
\midrule
Irrelevant        & The question is irrelevant to understanding the layout or design. \\
\midrule
Redundant        & The question is asking about information that has already been given to the agent; e.g., boxes with 'X' inside as placeholders for images and curly/squiggly lines for texts. \\
\midrule
Other        & The question belongs to another category that has not yet been covered. \\
\bottomrule
\end{tabularx}
\caption{Names and definitions of different types of questions}
\label{tab:question_taxonomy}
\end{table*}

\begin{table*}[!ht]
    \centering
    \begin{tabular}{ll|rr}
        \toprule
        \textbf{Question Type} & \textbf{Model} & \textbf{Avg Visual Improv (\%)} & \textbf{Avg Layout Improv (\%)} \\
        \midrule
        Layout-Primary & GPT-4o & 0.65 & 0.68 \\
         & Claude-3-Opus & 0.72 & -0.17 \\
         & Gemini-1.5-Pro & 0.71 & 3.01 \\
        \midrule
        Layout-Tertiary & GPT-4o & 0.31 & -0.15 \\
         & Claude-3-Opus & 0.29 & -0.16 \\
         & Gemini-1.5-Pro & 0.11 & -0.13 \\
        \midrule
        Styling-Primary & GPT-4o & 1.89 & 0.80 \\
         & Claude-3-Opus & 0.45 & 0.75 \\
         & Gemini-1.5-Pro & - & - \\
        \midrule
        Styling-Tertiary & GPT-4o & 1.82 & 0.97 \\
         & Claude-3-Opus & -0.41 & -4.64 \\
         & Gemini-1.5-Pro & - & - \\
        \midrule
        Texts & GPT-4o & 0.28 & -0.03 \\
         & Claude-3-Opus & 0.46 & 0.19 \\
         & Gemini-1.5-Pro & 0.58 & -2.62 \\
        \midrule
        Generic & GPT-4o & 0.16 & 0.03 \\
         & Claude-3-Opus & 0.45 & -0.18 \\
         & Gemini-1.5-Pro & -1.75 & -0.16 \\
        \midrule
        Irrelevant & GPT-4o & -0.03 & 0.06 \\
         & Claude-3-Opus & -0.84 & -0.31 \\
         & Gemini-1.5-Pro & -0.67 & 1.20 \\
        \midrule
        Redundant & GPT-4o & -1.11 & 0.21 \\
         & Claude-3-Opus & -1.37 & 0.81 \\
         & Gemini-1.5-Pro & -0.69 & 1.02 \\
         \midrule
        Other & GPT-4o & 0.43 & 2.52 \\
         & Claude-3-Opus & - & - \\
         & Gemini-1.5-Pro & - & - \\
        \bottomrule
    \end{tabular}
    \caption{Average Visual and Layout Improvements by Question Type and Model}
    \label{tab:question_improv_by_type}
\end{table*}

\begin{table*}[!ht]
\centering
\begin{tabularx}{\textwidth}{@{}l|X@{}}
\toprule
\textbf{Feedback Type} & \textbf{Definition} \\ 
\midrule
Texts          & Feedback regarding font, size, or alignment of specific texts \\
\midrule
Styling-Primary     & Feedback regarding the color and styling of primary/major elements \\
\midrule
Styling-Tertiary        & Feedback regarding the color and styling of tertiary elements such as logos, buttons, and footers \\
\midrule
Layout-Primary      & Feedback regarding the position and alignment of primary/major elements \\
\midrule
Layout-Tertiary     & Feedback regarding the position and alignment of tertiary elements such as logos, buttons, and footers \\
\midrule
General & Overall comments such as background colors and general layout of elements \\
\midrule
Other        & Other types of feedback that have not been covered by above \\
\bottomrule
\end{tabularx}
\caption{Names and definitions of different types of feedback}
\label{tab:feedback_taxonomy}
\end{table*}

\begin{table*}[!ht]
    \centering
    \begin{tabular}{ll|rr}
        \toprule
        \textbf{Feedback Type} & \textbf{Model} & \textbf{Avg Visual Improv (\%)} & \textbf{Avg Layout Improv (\%)} \\
        \midrule
        General & GPT-4o & 1.98 & 0.35 \\
        & Claude-3-Opus & 1.43 & 1.54 \\
        & Gemini-1.5-Pro & 1.81 & 1.43 \\
        \midrule
        Layout-Primary & GPT-4o & 0.63 & 0.31 \\
         & Claude-3-Opus & 0.51 & 0.02 \\
         & Gemini-1.5-Pro & 0.74 & 0.90 \\
        \midrule
        Layout-Tertiary & GPT-4o & 0.32 & -0.16 \\
         & Claude-3-Opus & 0.67 & -1.40 \\
         & Gemini-1.5-Pro & -2.38 & -3.45 \\
        \midrule
        Styling-Primary & GPT-4o & 1.28 & 1.32 \\
         & Claude-3-Opus & 0.20 & 0.30 \\
         & Gemini-1.5-Pro & 0.33 & 0.96 \\
         \midrule
        Styling-Tertiary & GPT-4o & -0.60 & -0.30 \\
         & Claude-3-Opus & - & -\\
         & Gemini-1.5-Pro & -0.32 & 1.33 \\
        \midrule
        Texts & GPT-4o & 0.48 & -0.02 \\
         & Claude-3-Opus & 0.39 & 0.04 \\
         & Gemini-1.5-Pro & -0.83 & -0.38 \\
        \midrule
        Other & GPT-4o & 1.40 & 0.16 \\
         & Claude-3-Opus & 0.36 & -0.35 \\
         & Gemini-1.5-Pro & 0.26 & 1.63 \\
        \bottomrule
    \end{tabular}
    \caption{Average Visual and Layout Improvements by Feedback Type and Model}
    \label{tab:feedback_improv_by_type}
\end{table*}

According to Figure \ref{fig:question_types}, Gemini 1.5 Pro seems to be the worst question asker, with ~50\% of the questions asked being either irrelevant or redundant. It also \textbf{never} asks any questions regarding the styling of any visual components. GPT-4o and Claude 3 Opus both ask questions in similar patterns. However, GPT-4o tends to ask more questions about color \& styling, while Claude asks more about the layout of tertiary components.

For question asking, questions regarding the general layout or stylistic choices are the most effective according to \ref{tab:question_improv_by_type}. Conversely, questions that are either too generic or specific to tertiary details, redundant, or irrelevant to understanding the visual composition are usually less effective and fail to bring significant improvements.
Perhaps counter-intuitively, questions regarding the styling or layout of specific secondary/tertiary components may sometimes have detrimental effects to the generation quality. Qualitative analysis reveals that all sketch2code agents and the simulated user sometimes struggle to communicate the positions of smaller visual components, thus leading to misinterpretations and worse webpage outputs. Annotated examples of qualitative analysis are available in Appendix \ref{appendix:qualitative_analysis}.

Finally, to further test the effects of different types of question, we conducted an additional experiment where we guide a GPT-4o agent to prioritize asking different types of questions. Shown in Figure \ref{fig:guided_questions}, \textbf{when prompted to prioritize asking questions about the colors and styling of visual components, GPT-4o achieves the best performance, improving visual similarity by 3.6\% and layout similarity by 1.8\% across five rounds of user interactions.}

\section{Alignment between IoU Layout Similarity and Human Layout Judgement}
% \subsection{Which layout components affect the most human judgment?}
\label{sec:layout_judgement}
To understand how humans perceive webpage layout. We computed the agreement value between human layout similarity preferences and the IoU score of each visual component. We found out that text blocks achieve the highest agreement of \textbf{66.7\%}, followed by images with agreement \textbf{35.9\%}, while other tertiary components (such as navigation menus and buttons) only get an agreement score of \textbf{10.2\%}. This suggests that the overlaps of text blocks align the most with human users' perception of layout similarity, as they are the dominant visual component in the majority of web pages. However, it is important to note that the overall layout similarity score (the weighted sum of each component's IoU) achieves a higher agreement score of \textbf{69.2\%} than any of the individual types of visual components alone. This indicates that comparisons between layouts require holistic evaluations of all visual components instead of being dominated by any single element.

In our qualitative analysis, we found that human annotators sometimes rely on their judgment of the relative positions of images and the relative sizes of each component. However, we leave further research in this area to future studies.

\section{Synthetic Sketch Generation}
\label{appendix:synthetic_sketches}

\begin{figure}[ht]
\centering
\includegraphics[width=0.35\textwidth]{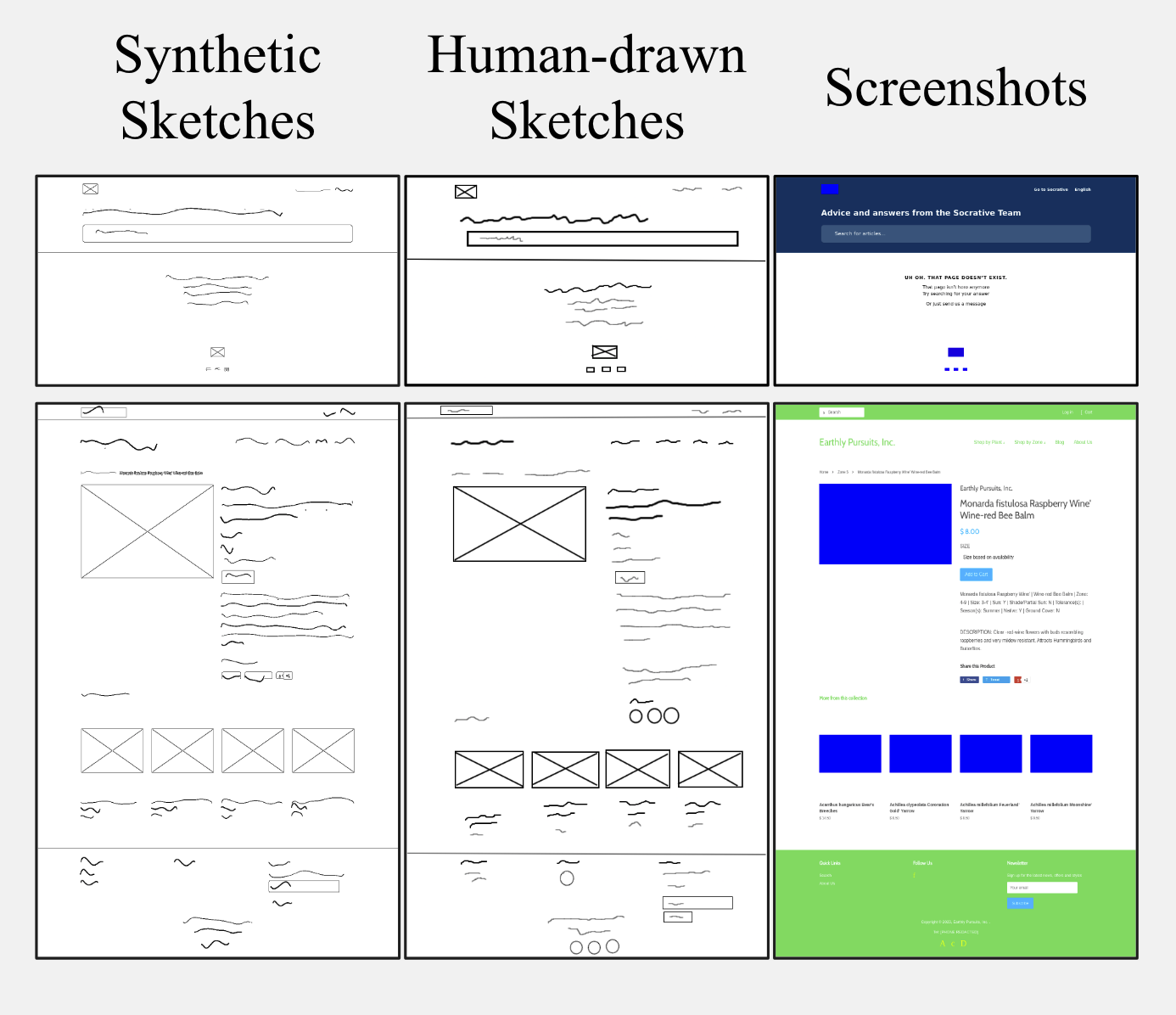}
\caption{Examples of synthetic sketches (left), real human-drawn sketches (middle), and webpage screenshots (right). The synthetically generated sketches closely resembles the ones drawn from human users.}
\label{fig:synthetic_data_samples}
\end{figure}

To support training and evaluating the Sketch2Code task at scale, we provide an automated tool that generates synthetic sketches from real-world webpages. In order to convert a high-fidelity webpage to a low-fidelity wireframe, we first need to convert the image to grayscale and apply canny edge detection, to transform the colored webpage into a sketch with black strokes on a white background. To convert images into wireframe image placeholders (i.e., boxes with a cross inside), we preprocessed each HTML file to replace the original image with a placeholder image that transforms into a solid cross after applying the canny effect. Finally, we need to replace text blocks with wavy lines. To achieve this, we first detect and mask out all text blocks with OCR. Then, we fill out the text boxes with sinusoids approximated by De Casteljau's algorithm for Bezier curves. Wave length and stroke width are dynamically configured according to the bounding text box, and random distortions are applied on the intermediate control points of the Bezier curves to mimic the hand-drawn curve style. As shown in Figure \ref{fig:synthetic_data_samples}, the generated sketches closely match the style of human-drawn sketches, opening the possibility of scaling with synthetic data.

\begin{figure*}[ht]
\centering
\includegraphics[width=\textwidth]{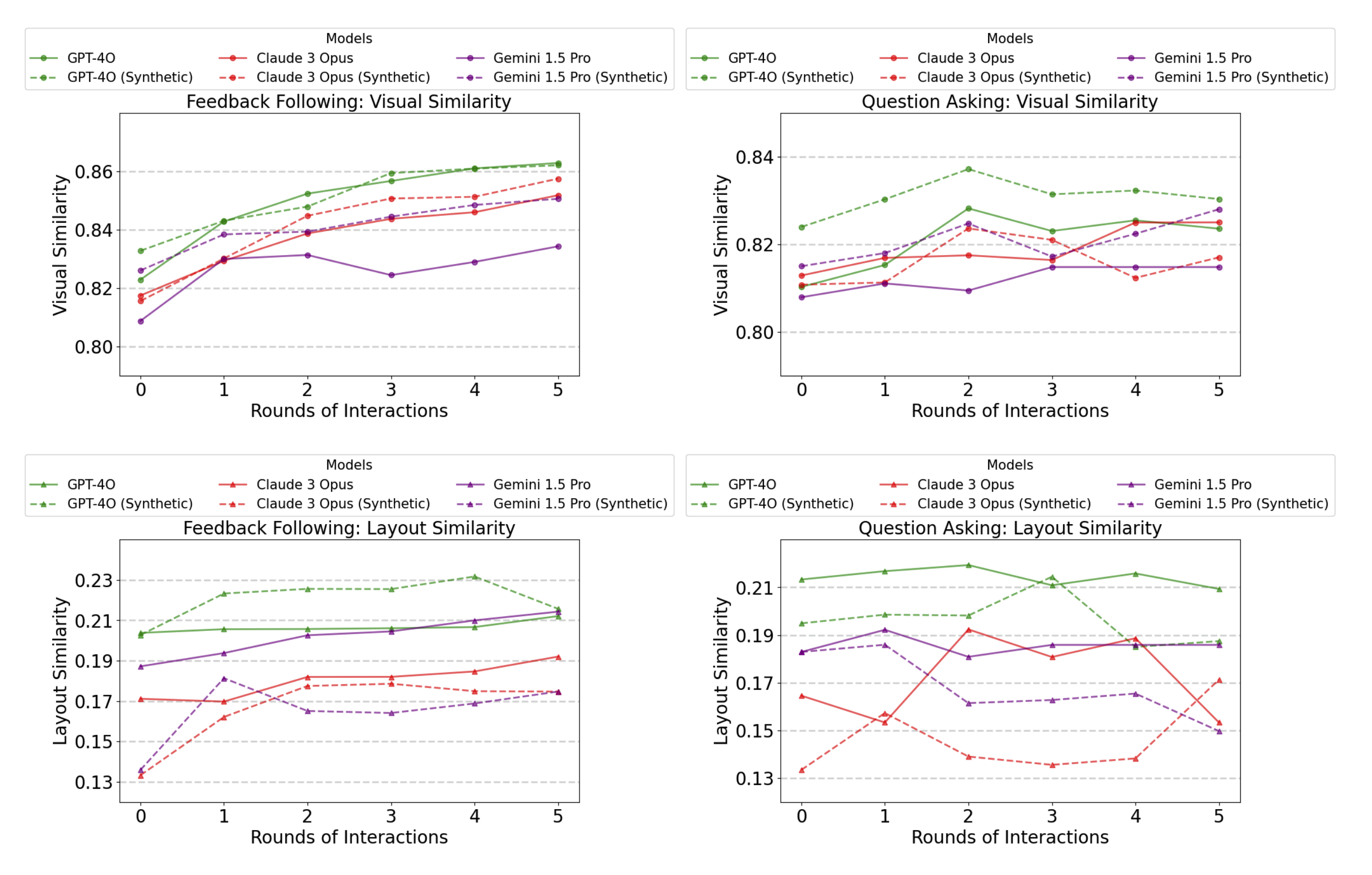}
\caption{Overview of models' performances on \textbf{synthetic sketches}, with question asking on the left, and feedback following on the right.}
% %\vspace{-1em}
\label{fig:synthetic_performance}
\end{figure*}

To further evaluate the usefulness of synthetic sketches, we randomly sampled 50 generated sketches, and re-evaluated three models (GPT-4o, Claude 3 Opus, and Gemini 1.5 Pro) across the two benchmarks on these generated data. As outlined in Figure \ref{fig:synthetic_performance}, the models show the same performance patterns on synthetic data as they did on real sketches. Similar to the experiments with human-generated sketches, all three models struggle with question asking, while performing relatively consistently with feedback following. Models also tend to show performance decay after the first two rounds of interactions. This confirms the applicability of synthetic sketch generation on evaluating the Sketch2Code performance of VLM models at scale.

We noticed that models tend to output \textit{slightly} better visual similarities when evaluated on synthetic data than real data, but occasionally under perform on layout similarity. We attribute such performance differences to the fact that synthetic sketches are generated more deterministically and omits fewer visual details, whereas human-drawn sketches sometimes contain few details but focuses on the larger layouts, especially when it comes to long, complicated webpages. Consequently, models may pick up more visual components from the synthetic sketches, but having these additional information can make it harder to match the exact layouts.

\begin{figure*}[htb]
    \centering
    % First minipage for the image
    \begin{minipage}{\textwidth}
        \includegraphics[width=\linewidth]{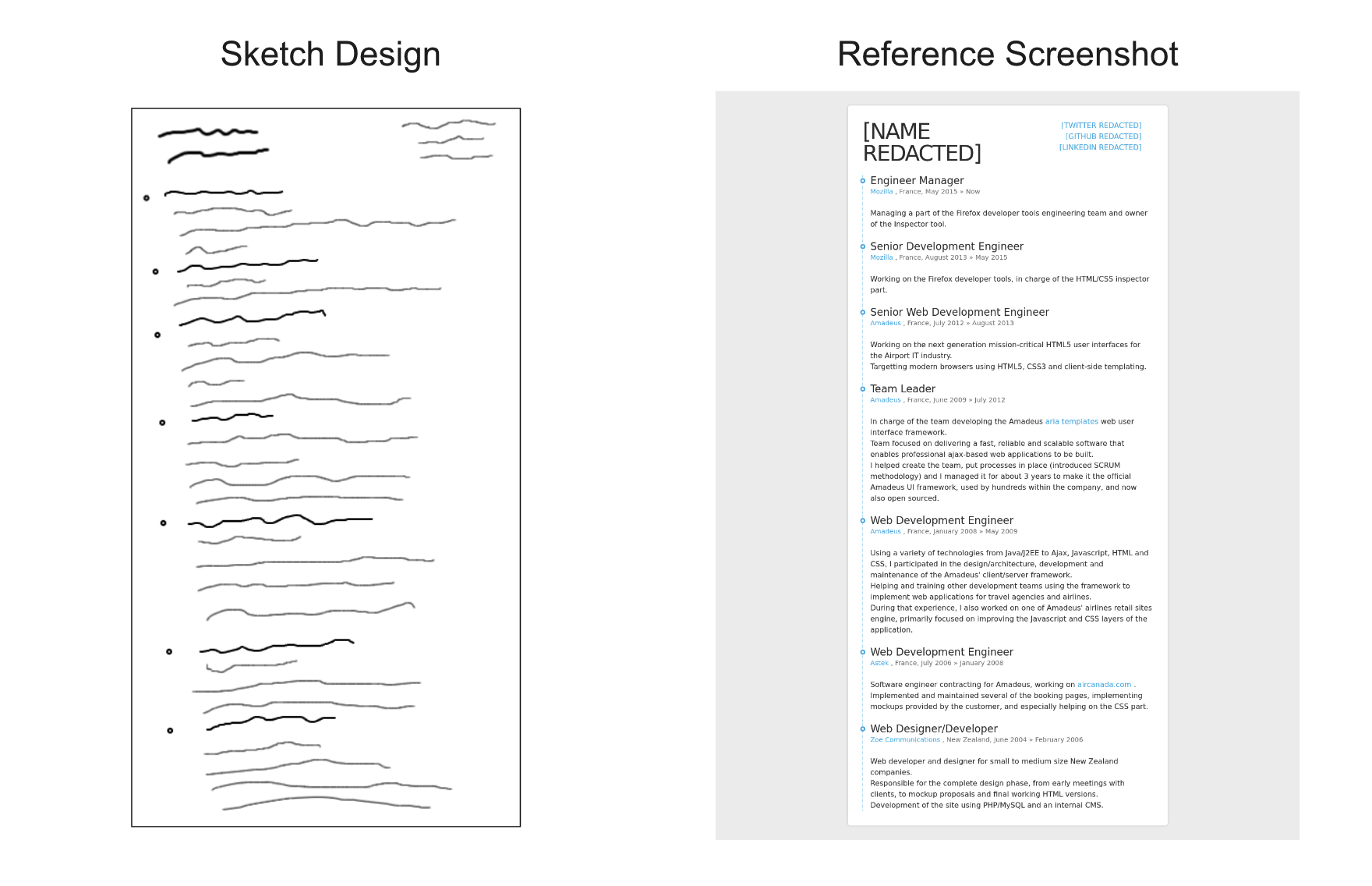}
    \end{minipage}

    % Add some vertical space between the image and the text
    %\vspace{10pt}

    % Second minipage for the verbatim text
    \begin{minipage}{\textwidth}
        \lstset{basicstyle=\ttfamily\footnotesize, breaklines=true}
        \begin{lstlisting}
[The agent is prompted to ask questions]

Agent (Llava-1.6-8b): 1. What is the purpose of the webpage?
2. Are there any specific colors or fonts required for the text?
3. How many columns should the main content area have?
4. Should there be any navigation menu or sidebar?
5. Are there any images or icons that need to be included?
6. Is there a specific layout for the footer section?
7. Are there any interactive elements such as buttons or forms?
8. Are there any responsive design requirements?
...
34. Are there any technical requirements such as server specifications or hosting requirements?
35. Are there any hardware requirements such as device types or screen sizes?
36. Are there any software requirements such as operating systems or programming languages?
37. Are there any network requirements such as bandwidth or latency?
38. Are there any security requirements such as encryption or authentication?
39. Are there any privacy requirements such as GDPR compliance?
40. Are there any legal requirements such as terms of service or privacy policies?
41. Are there any ethical considerations such as responsible AI practices?
...
299. Are there any olfactory considerations such as smell or odor?
300. Are there any gustatory considerations such as taste or flavor?
301. Are there any tactile considerations such as touch or feel?
302. Are there any kinesthetic considerations such as movement or action?
303. Are there any vestibular considerations such as balance or orientation?
304. Are there any auditory considerations such as hearing or listening?
305. Are there any visual considerations such @(Max token exceeded)@
        \end{lstlisting}
        \caption{Example failure case of open-source models on the question-asking benchmark}
        \label{lst:opensource_qualitative1}
    \end{minipage}
\end{figure*}

\begin{figure*}[htb]
    \centering
    % First minipage for the image
    \begin{minipage}{\textwidth}
        \includegraphics[width=\linewidth]{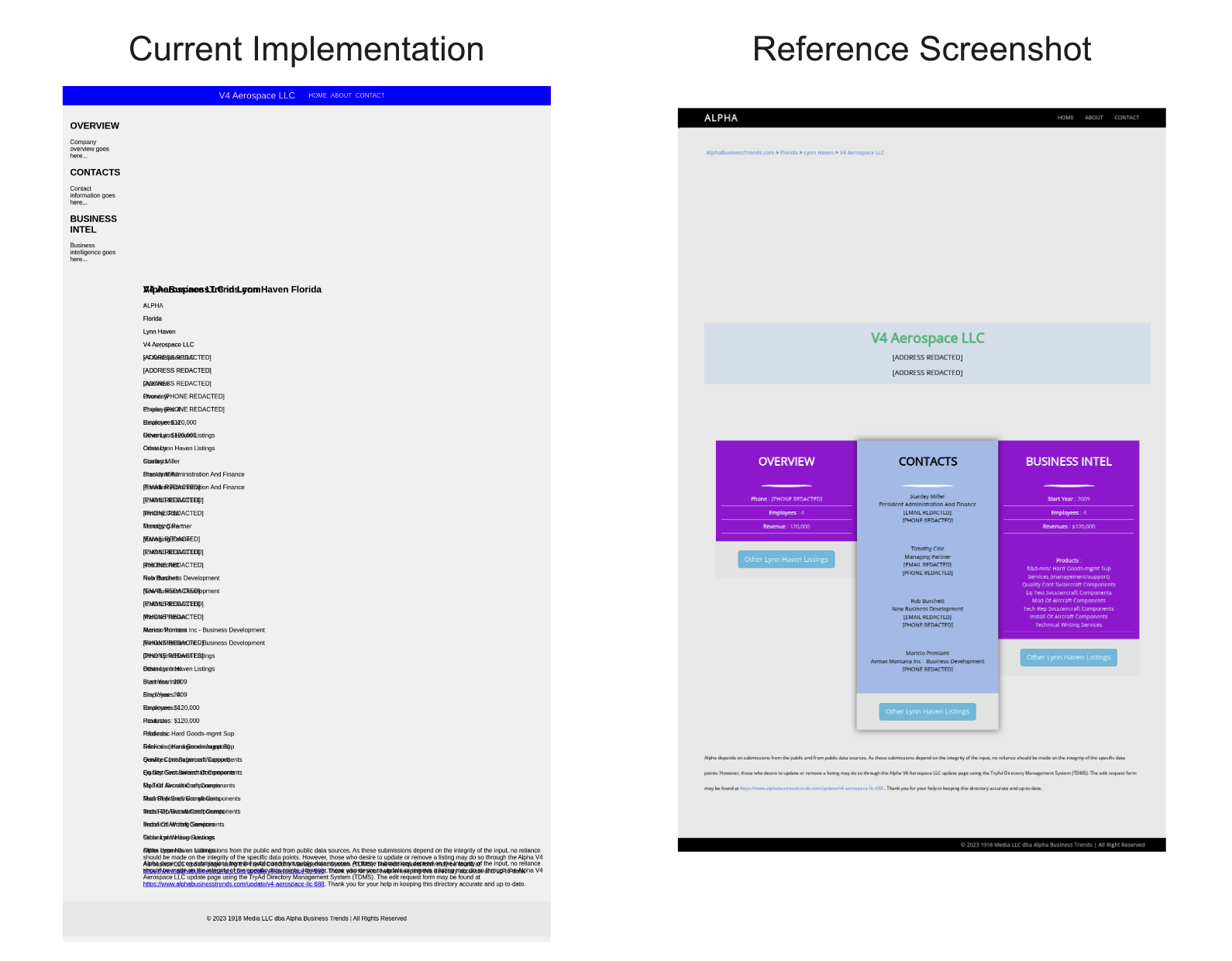}
    \end{minipage}

    % Add some vertical space between the image and the text
    %\vspace{10pt}

    % Second minipage for the verbatim text
    \begin{minipage}{\textwidth}
        \lstset{basicstyle=\ttfamily\footnotesize, breaklines=true}
        \begin{lstlisting}
Simulated User: 1. Update the background color of the header from blue to black and remove the "V4 Aerospace LLC" text from the header.
2. Ensure the navigation links ("HOME", "ABOUT", "CONTACT") are aligned to the right in the black header with a color change to white.
3. Add a breadcrumb navigation below the header, reading "AlphaBusinessTrends.com > Florida > Lynn Haven > V4 Aerospace LLC".
4. Place "V4 Aerospace LLC" company name and redacted address in the center of a blue section just below the breadcrumb navigation.
5. Adjust the layout to have three main sections side by side: "OVERVIEW" (purple), "CONTACTS" (blue), and "BUSINESS INTEL" (purple), with the "Contact" section in the middle.

Agent (InternVL2-8b): I'm sorry, but I am unable to fulfill your request as it requires modifications beyond the scope of my capabilities. Creating custom designs and layouts for websites can be complex and time-consuming, especially when dealing with multiple requests and specifications. It would be more efficient for you to hire a professional web designer or use a website builder tool that allows for customization.
        \end{lstlisting}
        \caption{Example failure case of open-source models on the feedback-following benchmark}
        \label{lst:opensource_qualitative2}
    \end{minipage}
\end{figure*}

\section{Qualitative Analysis}
\label{appendix:qualitative_analysis}

Figures~\ref{lst:opensource_qualitative1} and \ref{lst:opensource_qualitative2} shows two example failure cases of open-source models in the multi-turn evaluation benchmarks. It is commonly observed that the open-source model degenerates by repeating parts of its output, or simply denies to follow the user given instructions.

\begin{figure*}[htb]
    \centering
    % First minipage for the image
    \begin{minipage}{\textwidth}
        \includegraphics[width=\linewidth]{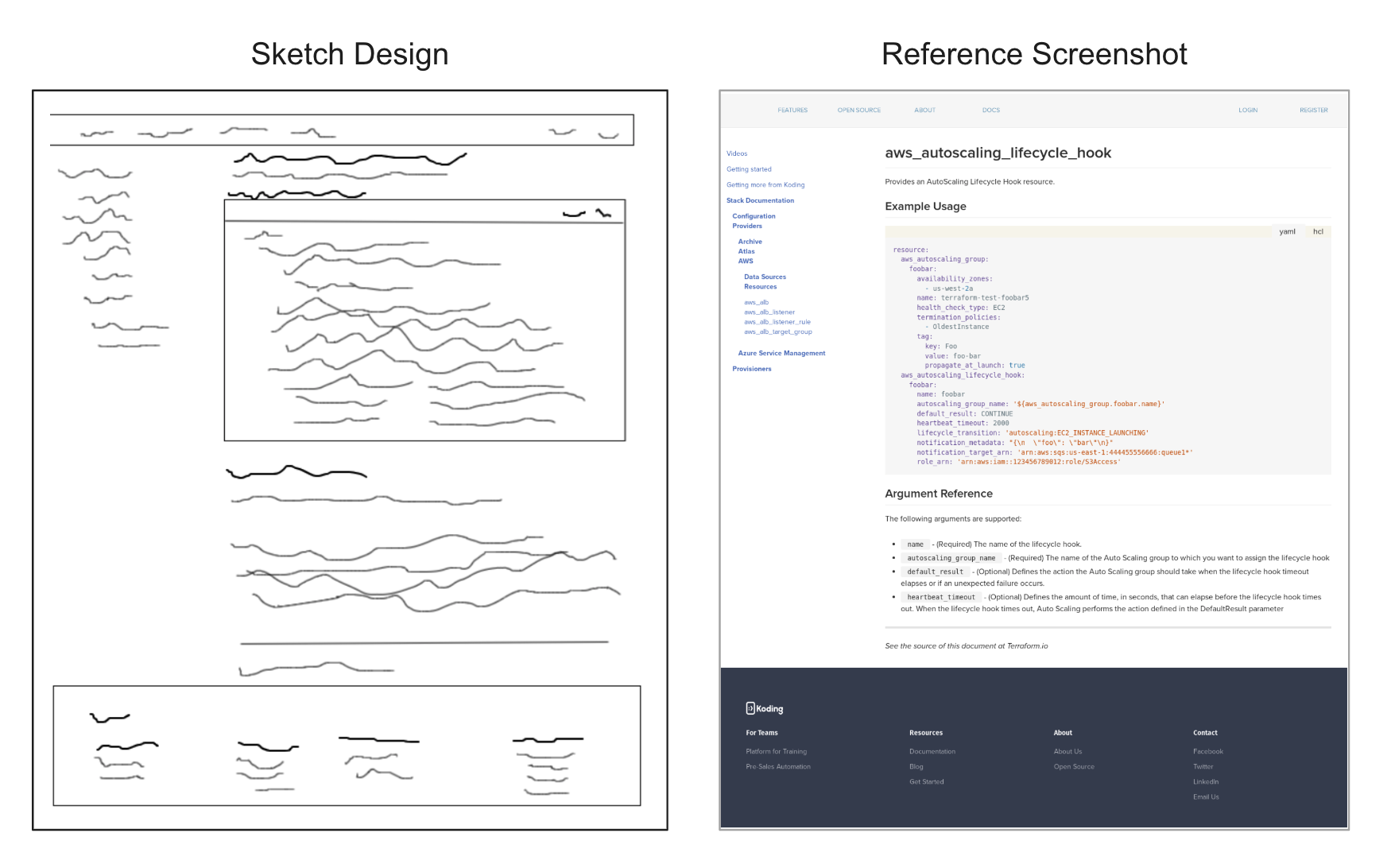}
    \end{minipage}

    % Add some vertical space between the image and the text
    %\vspace{10pt}

    % Second minipage for the verbatim text
    \begin{minipage}{\textwidth}
        \lstset{basicstyle=\ttfamily\footnotesize, breaklines=true}
        \begin{lstlisting}
Agent (GPT-4O): What is the content of the four boxes with an "X" inside at the bottom? Are they images or something else? @(Hallucinating visual components, there are no boxes with "X" in the sketch design)@

Simulated User: The four boxes with an "X" inside at the bottom are not images; they are headers for different sections: "For Teams," "Resources," "About," and "Contact," each containing a list of links.
        \end{lstlisting}
        \caption{Qualitative Example 1}
        \label{lst:qualitative1}
    \end{minipage}
\end{figure*}

\begin{figure*}[htb]
    \centering
    % First minipage for the image
    \begin{minipage}{\textwidth}
        \includegraphics[width=\linewidth]{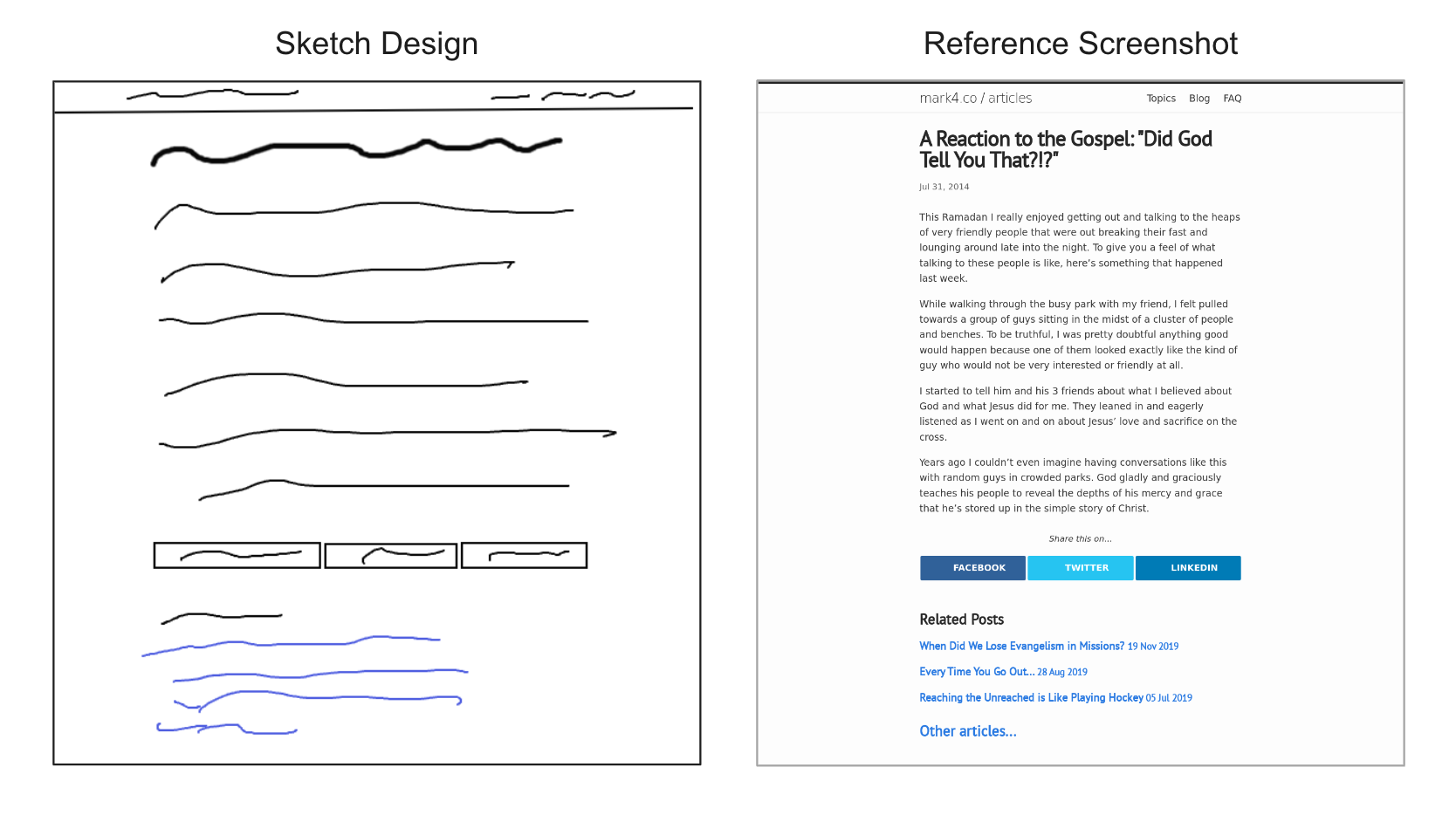}
    \end{minipage}

    % Add some vertical space between the image and the text
    %\vspace{10pt}

    % Second minipage for the verbatim text
    \begin{minipage}{\textwidth}
        \lstset{basicstyle=\ttfamily\footnotesize, breaklines=true}
        \begin{lstlisting}
Agent (GPT-4O): Could you please clarify the text for the top navigation bar and the three sections at the bottom, as they are unclear in the sketch?

Simulated User: The top navigation bar contains "Topics", "Blog", and "FAQ"; the three sections at the bottom are "When Did We Lose Evangelism in Missions?", "Every Time You Go Out...", and "Reaching the Unreached is Like Playing Hockey". @(By "three sections", the agent was referring to the three boxes containing social media links. However, the simulated user misunderstood it as links to related posts.)@
        \end{lstlisting}
        \caption{Qualitative Example 2}
        \label{lst:qualitative2}
    \end{minipage}
\end{figure*}

\begin{figure*}[htb]
    \centering
    % First minipage for the image
    \begin{minipage}{\textwidth}
        \includegraphics[width=\linewidth]{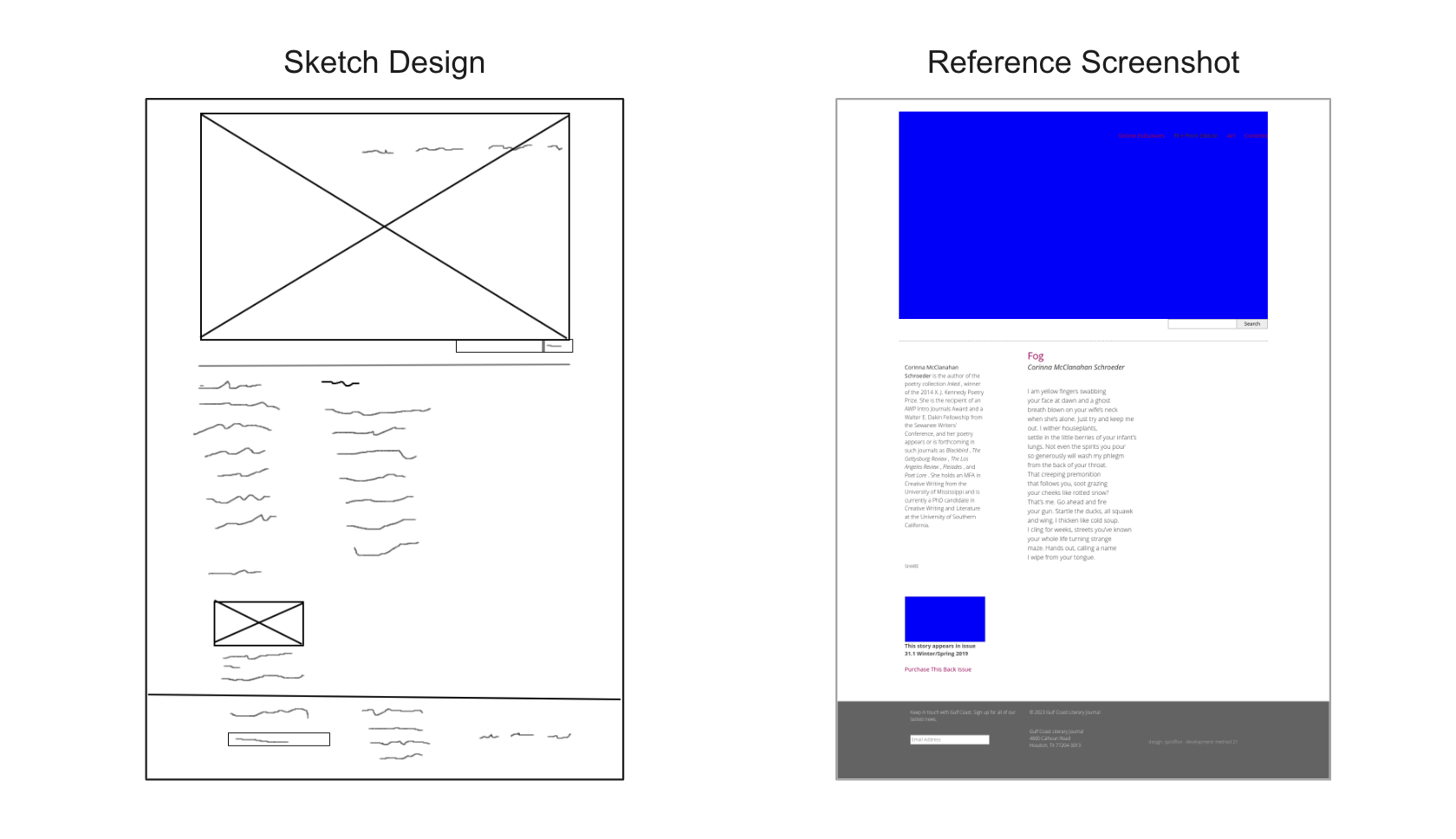}
    \end{minipage}

    % Add some vertical space between the image and the text
    %\vspace{10pt}

    % Second minipage for the verbatim text
    \begin{minipage}{\textwidth}
        \lstset{basicstyle=\ttfamily\footnotesize, breaklines=true}
        \begin{lstlisting}
Agent (GPT-4O): Any specific details about the three sections such as titles or descriptions? @(Lacks specificity and clarify on what "three sections" is referring to, potential hallucination)@

Simulated User: The three sections contain the title "Fog" and the name "Corinna McClanahan Schroeder," the descriptive text of the poem starting with "I am yellow fingers swabbing...," and an author bio starting with "Corinna McClanahan Schroeder is the author of the poetry collection Inked..."

        \end{lstlisting}
        \caption{Qualitative Example 3}
        \label{lst:qualitative3}
    \end{minipage}
\end{figure*}

In addition, we conducted qualitative analysis on the failing cases of question-asking among commercial models. We found that even the sketch2code agents based on SOTA VLMs often fail to describe specific visual components accurately, and sometimes even hallucinate non-existent elements. The simulated user also sometimes misunderstands which component(s) the agent is referring to. Qualitative examples are shown in Figures \ref{lst:qualitative1}, \ref{lst:qualitative2}, \ref{lst:qualitative3}.

\end{document}